\documentclass[lettersize,journal]{IEEEtran}
\usepackage{amsmath,amsfonts}
\usepackage{algorithmic}
\usepackage{algorithm}
\usepackage{array}
\usepackage{textcomp}
\usepackage{stfloats}
\usepackage{url}
\usepackage{verbatim}
\usepackage{graphicx}
\usepackage{cite}
\usepackage{booktabs}  % Add this line to include the booktabs package
\usepackage{multirow}
\usepackage{caption}
\usepackage{subcaption}
\usepackage{csquotes}

\usepackage{xcolor}
\usepackage{fancyhdr}
\usepackage{wrapfig}
\usepackage{array}
\usepackage{lipsum}
\usepackage{pifont}

\newcommand{\sect}[1]{Section~\ref{#1}}

\begin{document}
\bstctlcite{IEEEexample:BSTcontrol}
\title{PUL-Inter-slice Defender: An Anomaly Detection Solution for Distributed Slice Mobility Attacks}

%\author{IEEE Publication Technology,~\IEEEmembership{Staff,~IEEE,}
        % <-this % stops a space
%\thanks{This paper was produced by the IEEE Publication Technology Group. They are in Piscataway, NJ.}% <-this % stops a space
%\thanks{Manuscript received April 19, 2021; revised August 16, 2021.}
%}

\author{\IEEEauthorblockN{Ricardo Misael~Ayala Molina\textsuperscript{1}%, Nathalie Wehbe\textsuperscript{1}
, Hyame Assem Alameddine\textsuperscript{2}, Makan Pourzandi\textsuperscript{2}, Chadi Assi\textsuperscript{1}}  \\\textit{\textsuperscript{1}Concordia University, Montreal, Canada}  \textit{\textsuperscript{2}Ericsson, Montreal, Canada}}%\\\{ricardo.ayalamolina, nathalie.wehbe, chadi.assi\}@concordia.ca, \\  \{hyame.a.alameddine, makan.pourzandi\}@ericsson.com  }

% The paper headers
%\markboth{Journal of \LaTeX\ Class Files,~Vol.~14, No.~8, August~2021}%
%{Shell \MakeLowercase{\textit{et al.}}: A Sample Article Using IEEEtran.cls for IEEE Journals}

%\IEEEpubid{0000--0000/00\$00.00~\copyright~2021 IEEE}
% Remember, if you use this you must call \IEEEpubidadjcol in the second
% column for its text to clear the IEEEpubid mark.

\maketitle

\begin{abstract}

\textcolor{black}{Network Slices (NSs) are virtual networks operating over a shared physical infrastructure, each designed to meet specific application requirements while maintaining consistent Quality of Service (QoS). In Fifth Generation (5G) networks, User Equipment (UE) can connect to and seamlessly switch between multiple NSs to access diverse services. However, this flexibility, known as Inter-Slice Switching (ISS), introduces a potential vulnerability that can be exploited to launch Distributed Slice Mobility (DSM) attacks, a form of Distributed Denial of Service (DDoS) attack. %targeting both NSs and the 5G control plane. To address this challenge, 
To secure 5G networks and their NSs against DSM attacks, we present in this work, PUL-Inter-Slice Defender; an anomaly detection solution that leverages Positive Unlabeled Learning (PUL) and incorporates a combination of Long Short-Term Memory Autoencoders and K-Means clustering. PUL-Inter-Slice Defender leverages the Third Generation Partnership Project (3GPP) key performance indicators and performance measurement counters as features for its machine learning models to detect DSM attack variants while maintaining robustness in the presence of contaminated training data. When evaluated on data collected from our 5G testbed based on the open-source free5GC and UERANSIM, a UE/ Radio Access Network (RAN) simulator; PUL-Inter-Slice Defender achieved F1-scores exceeding 98.50\% on training datasets with 10\% to 40\% attack contamination, consistently outperforming its counterpart Inter-Slice Defender and other PUL based solutions combining One-Class Support Vector Machine (OCSVM) with Random Forest and XGBoost.}

% We evaluate the framework on a 5G testbed using Free5GC and UERANSIM, comparing it against three benchmarks: the original Inter-Slice Defender, and two PUL-based models combining One-Class Support Vector Machine (OCSVM) with Random Forest and XGBoost, respectively. PUL-Inter-Slice Defender achieves F1-scores above 98.50\% on training datasets with 10\% to 40\% attack contamination, consistently outperforming all benchmarks. These results demonstrate its robustness and effectiveness in detecting DSM attacks under realistic and challenging conditions.}
\end{abstract}

\vspace{-0.5cm}
\section{Introduction}
%\vspace{-0.3cm}

Network slicing is a cornerstone technology and a vital enabler for Fifth Generation (5G) networks, allowing them to deliver services that accommodate a wide range of specifications including high data rates, ultra low latency, diverse security requirements, dense connectivity, among others \cite{chahbar2020comprehensive}. The critical importance of network slicing is demonstrated by its central role in the development of virtual end-to-end networks, designated as Network Slices (NSs), which are individually configured to serve specific vertical industry applications \cite{abbas2021network, chahbar2020comprehensive, zhang2019overview}.

Network slicing enables User Equipment (UE) to seamlessly transition between various NSs \cite{3GPP_process} to address specific operational requirements such as Quality of Service (QoS), security levels, and service cost, among other factors \cite{sajjad2022inter}. UEs can be smoothly reallocated from one NS to another either based on a choice they opted for (e.g., enhance their QoS, access another service, etc.) or based on the network state and configuration by the Mobile Network Operator (MNO) (e.g., based on NS load and management configuration, etc.) \cite{sajjad2022inter}.

The intrinsic versatility of NSs opens up new security risks, making them vulnerable to attacks \cite{olimid20205g,de2023survey}. Notably, each time a UE switches between NSs, a process known as Inter-Slice Switching (ISS) is launched \cite{sajjad2022inter}. ISS requires the UE to re-initiate the authentication and registration procedures among others \cite{3GPP_ueconnection}. These procedures generate a substantial increase in signaling traffic within the 5G Control Plane (CP) and between the CP and the UE. Malicious actors can induce numerous ISS events to cause a flood of signaling messages that can overload the 5G Network Functions (NFs) such as the Access Mobility Management Function (AMF) and the Session Management Function (SMF). This scenario, known as Distributed Slice Mobility (DSM) attack, can culminate in a Distributed Denial of Service (DDoS) attack and disrupts the connection for legitimate UEs \cite{sathi2020distributed}.

DSM attack along with its economical and performance damage was first discussed in \cite{sathi2020distributed}. The same authors then proposed a protocol to secure the network against DSM attacks \cite{sathi2021dsm}. The proposed protocol automatically selects an NS for a UE based on its subscription and services offered by external networks. Subsequently, \cite{bisht2023detection} proposed a method for detecting DSM attacks by evaluating the average waiting time and switching rate. Lately, we devised two variations of the DSM attack and created Inter-Slice Defender \cite{rayalam2024}, a binary classifier based on a Long Short-Term Memory (LSTM)-Autoencoder model to detect them, leveraging 3GPP Key Performance Indicators (KPI) and Performance Measurement (PM) counters. To the best of our knowledge, \cite{bisht2023detection} and \cite{rayalam2024} are the only works which addressed DSM attack detection while other works \cite{khan2022slicesecure,kuadey2021deepsecure,thantharate2020secure5g} focused on DDoS attack detection in general using Machine Learning (ML) techniques without evaluating the effectiveness of their solutions in detecting the DSM attack.

%Further, none of the aforementioned works discussed the robustness of their ML models in the presence of contamination which questions the effectiveness of their solutions in real world deployment. In fact, in practice, data collected for training purposes often includes noise, mislabeled instances, or data from different/overlapping distributions, which can significantly affect the accuracy and reliability of ML models. Addressing this challenge is essential for ensuring that the proposed solutions are robust and effective in operational environments. This highlights the need for further research into methods that can effectively handle contaminated datasets, ensuring that DSM attack detection models are resilient and efficient in real-world deployment.

\textcolor{black}{Furthermore, none of the aforementioned works examined the robustness of their ML models under contaminated training conditions, raising concerns about their effectiveness in real-world deployments. In practice, training data often contains noise, mislabeled instances, or samples from overlapping distributions, all of which can undermine model accuracy and reliability \cite{papivc2023conditional}. %\red{[HA: Add reference]}. 
Addressing this challenge is essential for developing DSM attack detection solutions that remain resilient and effective in operational environments, underscoring the need for further research into methods capable of handling contaminated datasets.}

% this is according dr. Chadi: \textcolor{red}{In practice, data collected for training purposes often includes noise, mislabeled instances, or data from different/overlapping distributions, which can significantly affect the accuracy and reliability of ML models. However, it is important to note that none of the aforementioned works considered the impact of contaminated training datasets in their models, despite this being a common situation in real-world deployments. Addressing this challenge is essential for ensuring that the proposed solutions are robust and effective in operational environments. This highlights the need for further research into methods that can effectively handle contaminated datasets, ensuring that DSM attack detection models are more resilient and applicable in real-world scenarios.}

%In order to fill this gap, this work builds upon our previous research on Inter-Slice Defender \cite{rayalam2024} that did not show robustness in the presence of contamination, and proposes a novel network-slicing anomaly detection solution that incorporates Positive-Unlabeled Learning (PUL) into the training process \cite{sevetlidis2024dense} to detect two variations of the DSM attack, mainly, the Random Slice Attack (RSA) and Target Slice Attack (TSA). RSA consists of switching UEs to random NSs while TSA accounts for switching UEs to pre-selected targeted NS.

\textcolor{black}{To address this gap, we build upon our previous work on Inter-Slice Defender \cite{rayalam2024}, which lacked robustness under such conditions, and propose a novel network-slicing anomaly detection solution that integrates Positive-Unlabeled Learning (PUL) into the training process \cite{sevetlidis2024dense}. Our model is designed to detect two DSM attack variants: Random Slice Attack (RSA), where UEs are switched to random NSs, and Target Slice Attack (TSA), where UEs are redirected to pre-selected NSs.}

\textcolor{black}{PUL assumes a set of labeled positive data and a set of unlabeled data that includes both positive (i.e., benign data in our work) and negative (i.e., DSM attack) samples, in order to build a classifier capable of distinguishing between them \cite{bekker2020learning}. This solution is well-suited for real-world anomaly detection, where benign data can be reliably identified by experts during normal network operations. The assumption of unlabeled data aligns with typical unsupervised learning scenarios involving contaminated datasets, where training data predominantly consists of benign samples but may include undetected attacks. Therefore, PUL provides a solid methodological foundation for improving the robustness of Inter-Slice Defender in detecting DSM attacks, leading to the development of our proposed PUL-Inter-Slice Defender.}

Our contributions are summarized as follows:
\begin{itemize}

\item %We develope PUL-Inter-Slice Defender, a novel network slicing anomaly detection solution that incorporates PUL to detect variations of the DSM attack in the presence of a training dataset comprising a blend of normal behavior samples and attack instances.Besides, the framework employs an LSTM-Autoencoder model to extract features from latent spaces, profiling ISS events through its related  5G processes such as UE authentication, registration, PDU session establishment, and deregistration. K-Means clustering is then used to group these features into clusters to distinguish between positive and negative samples.
\textcolor{black}{We develop PUL-Inter-Slice Defender, a novel network slicing anomaly detection solution that incorporates PUL to detect DSM attack variations in contaminated training datasets comprising a blend of normal behavior samples and attack instances. Our approach employs an LSTM-Autoencoder to extract latent features profiling ISS events across key 5G procedures, including UE authentication, registration, PDU session establishment, and deregistration. These features are then clustered using K-Means to differentiate between benign and attack patterns.}

\item %We train and test our PUL-Inter-Slice Defender model using a training dataset including both normal and attack samples \textcolor{blue}{in the context of 5G NSs}. To the best of our knowledge, we are the first to explore the efficiency of PUL in detecting network slicing attacks under these conditions.
\textcolor{black}{We train and test the PUL-Inter-Slice Defender using a dataset containing both benign and attack samples within a 5G network slicing environment. To the best of our knowledge, this is the first study to assess the effectiveness of PUL for detecting slicing-related attacks under such conditions.}

\item Using the open source free5GC testbed \cite{free5GC} and the UERANSIM \cite{UERANSIM_Aligungr} simulator, we build a 5G network with four different NSs and we adapt it to emulate different variations of the DSM attack. 

\item %We simulate normal and DSM attack network traffic on our 5G testbed and collect the related data. We leverage the collected data to calculate 3GPP KPIs and PM counters that are used by our PUL-Inter-Slice Defender solution to detect DSM attacks. To the best of our knowledge, the generated dataset is the first of its kind for network slicing anomaly detection.
\textcolor{black}{We emulate both normal and DSM attack traffic on our 5G testbed and extract the corresponding data to compute 3GPP KPIs and PM counters, which serve as input to the PUL-Inter-Slice Defender. To the best of our knowledge, the resulting dataset is the first of its kind for network slicing anomaly detection.}

\item \textcolor{black}{We developed two PUL-based solutions: one combining One-Class Support Vector Machine (OCSVM) with Random Forest (RF), hereafter called PUL-OCSVM-RF, and another combining OCSVM with XGBoost, subsequently termed PUL-OCSVM-XGBoost. We use these as benchmarks, along with the Inter-Slice Defender, to compare against PUL-Inter-Slice Defender.}

\item  \textcolor{black}{Our experimental results demonstrate PUL-Inter-Slice Defender robustness in detecting RSA and TSA, with an average F1-score exceeding 98\% across training datasets containing 10\% to 40\% attack samples. This performance substantially surpasses that of the benchmark models, including Inter-Slice Defender, PUL-OCSVM-RF, and PUL-OCSVM-XGBoost, none of which exceed an F1-score of 89.87\% under the same conditions.}

\end{itemize}
%This performance significantly outperforms its counterpart, Inter-Slice Defender, which records F1-score values between 53.11\% and 80.74\% at the same contamination levels.

%The remainder of the paper is organized as follows: Section \ref{sec:relatedWork} presents the literature review. Section  \ref{sec:DSMattack}  outlines two DSM attack variations, alongside their assumptions and threat model. Section \ref{sec:InterSliceDefender} provides an overview of Inter-Slice Defender that we extended in this work. Section \ref{sec:methodology} details the architecture of PUL-Inter-Slice Defender and outlines the PUL assumptions considered in its design. Section \ref{sec:environmentalSetup} introduces our simulation setup including our 5G testbed, simulation scenarios, and datasets. Section \ref{sec:DSM_onCP} highlights the impact of RSA and TSA on the 5G network. Section \ref{sec:expResults} discusses the experimental results. Section \ref{sec:deploy} proposes a deployment option for PUL-Inter-Slice Defender.Finally, the work is concluded in Section \ref{sec:conclussi}.

\textcolor{black}{The remainder of the paper is structured as follows: Section  \ref{sec:relatedWork} reviews related work. Section \ref{sec:DSMattack} describes the DSM attack variations and threat model, while Section \ref{sec:InterSliceDefender} summarizes the original Inter-Slice Defender. Section \ref{sec:methodology} presents the PUL-Inter-Slice Defender architecture and assumptions. Section \ref{sec:environmentalSetup} outlines the emulation setup and datasets. Section \ref{sec:DSM_onCP} examines the impact of RSA and TSA. Section \ref{sec:expResults} reports experimental results; Section \ref{sec:deploy}  discusses implementation considerations in real-world 5G environments, and Section \ref{sec:conclussi} concludes the paper.}

\vspace{-0.3cm}
\section{Literature review} \label{sec:relatedWork}
%\subsection{Network Slicing Security}
\begin{table*}[t]
\centering
\footnotesize
\caption{State-of-the-art on DSM attack detection and PUL (X = Not Supported, \checkmark = Supported, N/A=Not Applicable).}
\resizebox{\textwidth}{!}{%
\begin{tabular}{c p{2.9cm} c c c >{\centering\arraybackslash}p{2.9cm} c c}
\toprule
\multirow{2}{*}{} & \multirow{2}{*}{\textbf{Reference}} & \multirow{2}{*}{\textbf{Approach}} & \multirow{2}{*}{\textbf{ML Model(s)}} & \multirow{2}{*}{\shortstack{\textbf{Type of} \\ \textbf{Feature}}} & \multirow{2}{*}{\shortstack{\textbf{Robustness in the} \\ \textbf{Presence of} \\ \textbf{Contamination}}} & \multirow{2}{*}{\textcolor{black}{\textbf{5G}}} & \multirow{2}{*}{\textbf{F1-score}} \\
 &  &  &  &  &  &  &  \\ \\ \midrule

\multicolumn{8}{c}{\textcolor{black}{\textit{Approaches to detect DSM attacks (\cite{bisht2023detection,rayalam2024}) and DDoS attacks based on flow-based features (\cite{khan2022slicesecure,kuadey2021deepsecure,thantharate2020secure5g}) in 5G network without incorporating PUL}}} \\
\addlinespace
{} & Bisht et al. \cite{bisht2023detection} & Two metrics & N/A & N/A & X & \textbf{\checkmark}& 91\%–94\% \\ 
{} & R.M.A. Molina et al. \cite{rayalam2024} & DL Model & LSTM-Autoencoder & PM Counters, KPIs & X & \textbf{\checkmark}& 98.75\% \\ 
{} & Khan et al. \cite{khan2022slicesecure} & DL model & BiLSTM & Flow-based features & X & \textbf{\checkmark}& 99.99\% \\ 
{} & Kuadey et al. \cite{kuadey2021deepsecure} & DL model & LSTM & Flow-based features & X & \textbf{\checkmark}& 99.965\% \\ 
{} & Thantharate et al. \cite{thantharate2020secure5g} & DL model & Deep Neural Network & Flow-based features & X & \textbf{\checkmark}& N/A \\
%{} & Sattar et al. \cite{sattar2019towards} & Mathematical model & N/A & N/A & X & \textbf{\checkmark}& N/A \\

\midrule
\multicolumn{8}{c}{\textcolor{black}{\textit{Approaches to detect DDoS attacks using PUL in technologies other than 5G}}} \\
\addlinespace                   

{} & \textcolor{black}{G. Long et al.\cite{long2024punet}} & \textcolor{black}{PUL, DL Model} & \textcolor{black}{VRNN} & \textcolor{black}{Latent feature represen-} & \textbf{\checkmark} & \textcolor{black}{X} & \textcolor{black}{N/A} \\ 
{} &  &  &  & \textcolor{black}{tation from raw traffic} &  &  & \\ 
{} & \textcolor{black}{R. Dilworth et al. \cite{dilworth2024harnessing}} & \textcolor{black}{PUL, DL Model} & \textcolor{black}{XGBoost, RF} & \textcolor{black}{Flow-based features} & \textbf{\checkmark} & \textcolor{black}{X} & \textcolor{black}{40.30\%–99.05\%} \\ 
{} &  &  & \textcolor{black}{SVM, Naïve Bayes} &  &  &  & \\ 
{} & \textcolor{black}{Z. Fan et al. \cite{fan2023self}} & \textcolor{black}{PUL, DL Model} & \textcolor{black}{Binary classifier} & \textcolor{black}{Flow-based features} & \textbf{\checkmark} & \textcolor{black}{X} & \textcolor{black}{94.51\%} \\ 
{} & \textcolor{black}{S. Lv. et al. \cite{lv2020intrusion}} & \textcolor{black}{PUL, DL Model} & \textcolor{black}{Custom neural network} & \textcolor{black}{Flow-based features} & \textbf{\checkmark} & \textcolor{black}{X} & \textcolor{black}{88.71\%–99.65\%} \\

\midrule
\multicolumn{8}{c}{\textcolor{black}{\textit{Our approach uses PUL in 5G networks, leveraging NS-specific features and 5G-specific datasets to detect DSM attacks}}} \\
\addlinespace

{} & \textbf{This work} & \textbf{PUL, DL/ML models} & \textbf{LSTM-Autoencoder} & \textbf{PM counters} & \textbf{\checkmark} & \textbf{\checkmark} & \textbf{*98.50\%–99.33\%} \\ 
{} &  &  & \textbf{and K-means} & \textbf{and KPIs} &  &  & \\ 

\bottomrule
\end{tabular}%
}

\label{table:comparison}
%\vspace{0.1cm}
*F1-score values across variable contamination/negative sample proportions ranging from 10\% to 40\% in the training dataset.
%\vspace{-0.5cm}
\end{table*}

%\vspace{-0.2cm}
\subsection{DSM Attack} \label{sec:DMSattack_researches}
%\vspace{-0.3cm}
A limited number of research studies have been dedicated to analyzing and detecting DSM attacks. For instance, Sathi et al.\cite{sathi2020distributed} were the first to propose the DSM attack and theoretically study the economic and performance damage it can cause on the network. They note that such damage exceeds those resulting from other DDoS attacks. Sajjad et al. \cite{sajjad2022inter} detailed 3GPP-based procedures enabling UE mobility and its transition between NSs. They highlighted critical UE mobility challenges and suggested potential avenues for future research.
Bisht et al. \cite{bisht2023detection} illustrated the consequences of ISS events that result in a DDoS attack. They employed an algorithm based on two metrics, average waiting time and switching rate, to identify and block compromised UEs. From a different perspective, Sathi et al. \cite{sathi2021dsm} introduced a preventive approach through a novel NS selection protocol. The protocol suggests that the network chooses the best NS for the UE based on its subscription and data network services which contradicts 3GPP 5G standards that allow the UE to request the services of a specific NS \textcolor{black}{\cite {3ggpPortal}}.

The work tackling DSM attack is limited to theoretical studies and does not consider the practical implementation of this attack in a real 5G network. Further, DSM attack detection solutions are limited to \cite{bisht2023detection}, which followed a simplistic algorithmic approach based on two metrics that were not tested on a real 5G network slicing-based dataset, thus overlooking the complexity of this attack.
\vspace{-0.3cm}
\subsection{Traditional DDoS Attacks Targeting Network Slices} \label{sec:NSapproach}
DDoS attacks are known to be the most disruptive attacks in the realm of cybersecurity \cite{al2023ddos,bremler2024exploiting}. While DSM has been identified as a DDoS attack on NSs, DDoS flooding attacks, such as User Data Protocol (UDP) lag and Transmission Control Protocol (TCP) SYN, among others, have also gained attention. DDoS flooding attacks can be caused by exploiting vulnerabilities in the high number of UEs in 5G networks and targeting its NSs. To detect these attacks, Deep Learning (DL) and mathematical models have been proposed.

For example, Khan et al. \cite{khan2022slicesecure} used a bidirectional-LSTM model to detect DDoS attacks (i.e., UDP flooding and TCP SYN attacks) in two NSs. They evaluated the impact of these attacks by measuring the incurred bandwidth and latency. %and created a dataset from compromised and non-compromised UEs.
Similarly, the authors in \cite{kuadey2021deepsecure} devised a solution leveraging LSTM to identify DDoS attacks and detect if UEs' NS requests are either legitimate or  malicious.
Further, Thantharate et al. \cite{thantharate2020secure5g} developed Secure5G that employs a Convolutional Neural Network (CNN) for early DDoS attack detection by analyzing network traffic patterns collected from a 5G testbed. Secure5G detects anomalous UEs' requests targeting multiple NSs and redirects suspicious UEs to a quarantine NS. %Sattar et al. \cite{sattar2019towards} proposed an NS isolation mechanism as a mitigation measure against DDoS attacks using a mathematical model. They evaluated their model’s performance by measuring its impact on bandwidth, response time, and round-trip time.

Most of the aforementioned work on NS attack detection focuses on general DDoS attacks that are not NS specific. Despite the efforts presented in some of these works to test these attacks on a 5G testbed and use the generated dataset for DDoS attack detection, their anomaly detection models remain limited to flow-based features that do not capture any NS-specific characteristics. Moreover, these studies do not examine nor evaluate the performance of their detection solutions in detecting the DSM attack and its variations.

\textcolor{black}{It is important to emphasize that there are other studies \cite{narmadha2025improved,wei2023reconstruction,shaikh2019autoencoder} that employ autoencoders, LSTM, clustering algorithms, or combinations of these methods for attack detection in domains such as traditional Internet-based networks. While these studies may give the impression of similarity to our proposed approach, they neither incorporate 5G-specific features nor adopt PUL in their methodologies. These two aspects are central to our approach, and their absence makes those studies unsuitable for direct comparison.}

\vspace{-0.3cm}
\subsection{Positive-Unlabeled Learning in DDoS Attack Detection }

%\textcolor{blue}{PUL is a paradigm that needs to be explored in the context of 5G technology, offering a promising research direction for anomaly detection. However, in technologies other than 5G,} The concept of learning from positive and unlabeled data has gained increasing attention in the field of ML for anomaly detection applications due to its practical significance in deployment scenarios where obtaining fully clean or labeled datasets is infeasible.

\textcolor{black}{Although PUL has been widely studied in various domains, its potential in 5G technology remains underexplored, particularly for anomaly detection. Conversely, the concept of learning from positive and unlabeled data has gained significant attention in the machine learning community for anomaly detection applications in other technologies, due to its practical relevance in deployment scenarios where obtaining fully clean or comprehensively labeled datasets is infeasible.}

\textcolor{black}{For example, \cite{long2024punet} %proposed PUNet, a PUL-based semi-supervised anomaly detection model that addresses distribution shift and data imbalance using Variational Recurrent Neural Network(VRNN)-based features and reconstruction-loss-guided pseudo-labeling. %Trained on CTU-13 and DoH2020 datasets, it uses CatBoost for classification and effectively detects DDoS and encrypted traffic anomalies. 
proposed PUNet, a PUL-based semi-supervised anomaly detection model that addresses distribution shift and data imbalance using Variational Recurrent Neural Network (VRNN)-based features and reconstruction-loss-guided pseudo-labeling. It uses CatBoost for classification and effectively detects DDoS and encrypted traffic anomalies.
\cite{dilworth2024harnessing} applies PUL to cloud-based DDoS detection using the BCCC-cPacket-Cloud-DDoS-2024 dataset. It frames DDoS attacks as positive samples and evaluates multiple classifiers. The study highlights PUL’s effectiveness in handling limited labeled data for detecting DDoS attacks in cloud environments.\cite{fan2023self} %This work proposes SRPU, a PUL-based framework for malicious traffic detection under class imbalance. It combines self-paced learning and a reweighting strategy to gradually select confident samples and handle noisy data. %Evaluated on CIC-IDS2017, UNSW-NB15, and APT datasets, SRPU effectively detects various attack types, including DDoS, even with limited labeled samples.
This work proposes SRPU, a PUL-based solution for malicious traffic detection under class imbalance. It combines self-paced learning with a reweighting strategy to gradually select confident samples and manage noisy data. SRPU effectively detects various attack types, including DDoS, even with limited labeled samples.
\cite{lv2020intrusion} introduced a method that uses non-negative risk estimator %\red{(nnPU) [HA:Is this abbreviation correct? do we actually need it?]}
%reply: The paper detects cyberattacks or network intrusions in general (for example, the paper uses KDD-99, NSL-KDD to run experiments)
learning for intrusion detection %\red{[HA: it helps if you can mention what kind of attacks they detected rather than just saying intrusion detection]}
in network systems. This approach treats cyber-attacks as positive samples in a PUL scenario and utilizes a risk estimator to calculate binary classification loss, specifically enhanced with focal loss to address data imbalance issues. }

Given the demonstrated success of PUL in various anomaly detection applications, we adopt this approach in our methodology for detecting DSM attacks in 5G networks, as obtaining clean, fully labeled datasets is often impractical. 
\vspace{-0.3cm}
\subsection{Overall Comparison to Our Approach}
%\vspace{-0.1cm}
In this paper, we highlight the lack of previous works that detect DSM attacks while considering deploying their solutions in real-world scenarios where obtaining a clean dataset is not always feasible. We discuss several key limitations of the related works in literature and provide a comprehensive comparison in Table \ref{table:comparison} with our work against them.
For instance, the works presented in this table \cite{rayalam2024,bisht2023detection,khan2022slicesecure,kuadey2021deepsecure,thantharate2020secure5g} operate under the assumption that the training dataset is clean, meaning it exclusively contains samples from the class of interest without any noise or irrelevant data. This assumption simplifies the learning process, as the model can focus on distinguishing between well-defined classes without the interference of mislabeled or out-of-distribution samples. Nevertheless, in many real-world scenarios, this assumption might not always be valid. In practice, training datasets are often contaminated with a certain percentage of unwanted instances \cite{tian2023leveraging}, such as mislabeled data, samples from other classes, or outliers. These contaminants can distort the learned model, leading to reduced performance, biased predictions, or even incorrect conclusions. Furthermore, some studies \cite{rayalam2024,khan2022slicesecure,kuadey2021deepsecure} achieved high F1-scores over 98\% by proposing approaches that leverage DL models. However, these results are based on ideal, uncontaminated datasets. Consequently, the robustness and generalization of these works against data contamination remain significant gaps in existing research, underscoring the importance of developing models that can maintain high performance in real-world conditions.

\textcolor{black}{In contrast to the assumptions made by previous studies, \cite{long2024punet,dilworth2024harnessing,fan2023self,lv2020intrusion} acknowledge that obtaining clean, uncontaminated training datasets is rarely feasible in practical settings. These works incorporate the PUL paradigm into their anomaly detection approaches, enabling their models to learn from datasets composed of both labeled and unlabeled data. As shown in Table \ref{table:comparison}, their methods achieve promising F1-scores exceeding 94\%, detecting various types of attacks such as DDoS, thereby demonstrating the effectiveness of PUL in handling real-world data contamination. However, despite these strengths, their approaches remain limited in scope. First, their evaluations are not conducted on 5G-specific datasets nor within a 5G testbed environment, raising concerns about their applicability to current and future mobile network architectures. Second, similar to \cite{bisht2023detection,khan2022slicesecure,kuadey2021deepsecure,thantharate2020secure5g}, these studies do not consider NS-specific features or behaviors. As a result, they overlook critical NS-level dynamics that are essential for accurately detecting sophisticated slice-based attacks such as DSM. Furthermore, none of the studies \cite{long2024punet,dilworth2024harnessing,fan2023self,lv2020intrusion} examine or evaluate the effectiveness of their detection models against DSM attack variations like RSA and TSA. Consequently, their robustness and generalization capabilities within the context of 5G and NS-oriented threat landscapes remain unaddressed. This highlights the need for specialized detection mechanisms, such as the one we propose, that are not only designed for contaminated data but are also tailored to detect DSM attacks within realistic 5G network environments using NS-aware features and real testbed implementations.}
\textcolor{black}{We provide a practical and comprehensive approach that combines the PUL paradigm with an LSTM-Autoencoder integrated with the K-means algorithm for feature extraction and clustering. This architecture is specifically designed to capture the complex and evolving patterns of DSM attacks in the presence of contaminated data. Unlike prior studies, our solution is implemented and evaluated in a realistic 5G testbed that supports network slicing (Section VI-A), from which we collect traffic to compute 3GPP-based features (e.g., KPIs and PM counters) for profiling the behavior of NSs and core NFs. We also examine the impact of DSM attacks on CPU usage in the CP NFs, demonstrating their disruptive effect on 5G operations. This end-to-end methodology underscores the robustness and practical value of our proposed solution for DSM attack detection in real-world 5G deployments.}

\vspace{-0.3cm}
\section{DSM Attack - Threat Model and Variations} \label{sec:DSMattack}
\begin{figure*}[h]
    \centering
    \includegraphics[width=0.85\linewidth]{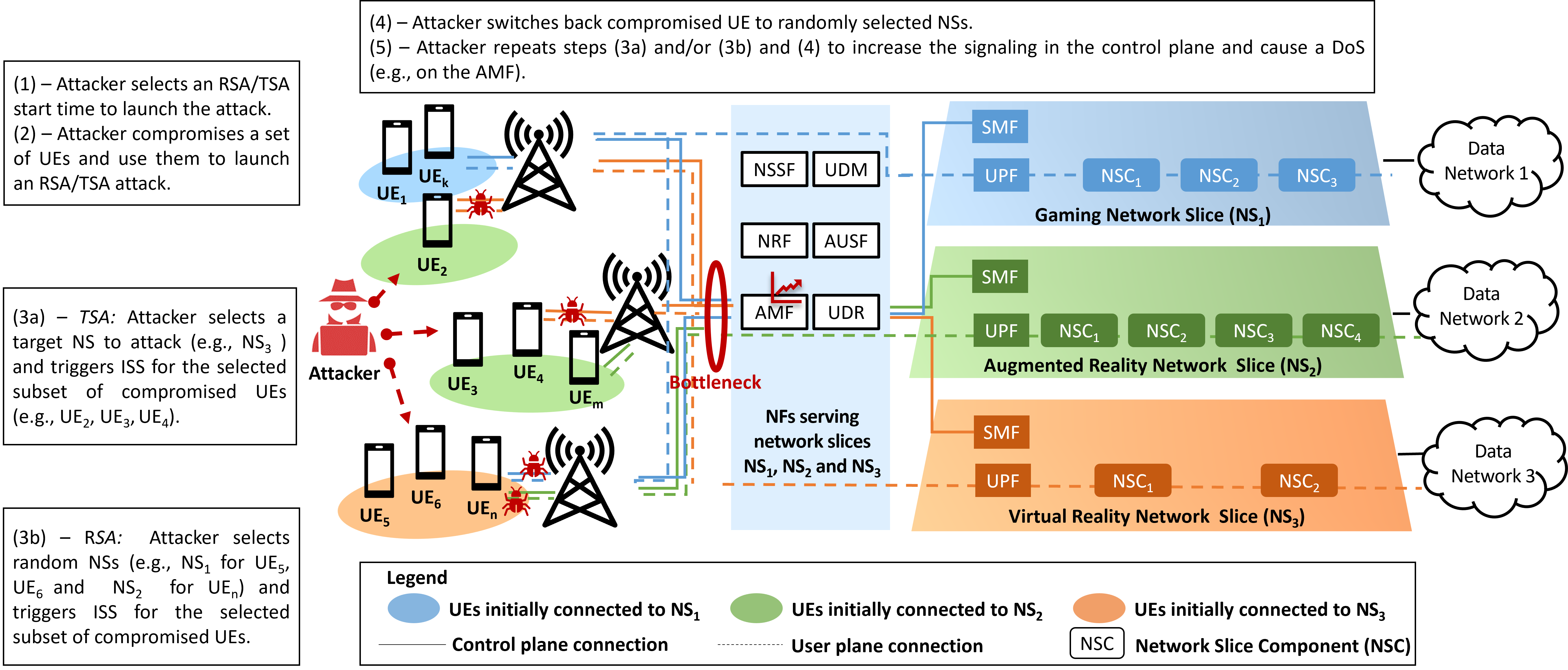}
    \captionsetup{justification=justified, singlelinecheck=false} 
    \caption{ \textcolor{black}{Overview of DSM attack variations, illustrating how compromised UEs initiate ISS to perform RSA and TSA, resulting in excessive signaling load on 5G Core NFs. The colored line indicates the targeted NSs.}}   
    \label{fig:RSA_and_TSA_attacks}
\end{figure*}

The DSM attack is a DDoS attack that exploits UEs ISS events to disrupt the performance of both the 5G network CP and its NSs. Such disruption is caused by the signaling associated with the high number of malicious ISS events. In fact, ISS triggers the PDU session release procedure to release the UE PDU session in its current NS, registration, and PDU session establishment to a new desired NS, and other procedures, such as UE authentication to the network \cite{sajjad2022inter,3GPP_ueconnection}. These procedures involve a sheer volume of signaling messages within the 5G CP and the NSs. In the following, we detail the DSM attack assumptions, and threat model while highlighting two of its variations (i.e., RSA and TSA). % that diverge from the approach presented in \cite{sathi2020distributed}.
%\vspace{-0.3cm}
\subsection{Attack Assumptions}
To perform RSA and TSA, these assumptions are followed:
%the following assumptions are made:
 \begin{enumerate}%[leftmargin=*]

\item{\textbf{Access to a set of compromised UEs.}} UEs and Internet of Things (IoT) devices are known to be highly vulnerable to attacks such as those noted in \cite{neshenko2019demystifying}. An attacker, can thus compromise a set of UEs and use them as a botnet to perform a DSM attack.

\item{\textbf{UEs NS configurations and credential information.}} We assume that the attacker has access to compromised UEs credentials (i.e., cryptographic keys, NSs configuration, etc.) and can use them to successfully connect to the MNO network and its NSs.

\item{\textbf{Remote activation of ISS.}} We assume that the attacker is able to remotely access and manipulate the compromised UEs in order to force them to trigger ISS to switch between their accessible NSs.

\end{enumerate}
%\vspace{-0.3cm}

\subsection{DSM Attack Variations - Threat Model}
In this work, we exploit two variations of the DSM attack, illustrated in Fig. \ref{fig:RSA_and_TSA_attacks}:
%\begin{itemize}

\textbf {Random Slice Attack (RSA).} To perform a RSA, the attacker selects different subsets of compromised UEs, and simultaneously switch them to randomly selected NSs. RSA creates varying loads on these random NSs as a result of the varying number of connected UEs. % will be connected to each of the random NSs.

\textbf {Target Slice Attack (TSA).} To perform a TSA, the attacker selects a target NS to attack and triggers ISS events to switch the compromised set of UEs to the selected NS.  

According to Fig. \ref{fig:RSA_and_TSA_attacks}, RSA and TSA can be performed as follows:

\begin{enumerate}%[leftmargin=*]
 \item{\textbf{Select RSA/TSA start time.}} The attacker identifies the network's peak times, during which a substantial number of UEs are connected to the network in order to perform the RSA/TSA. This will impose a more pronounced impact on the CP's NFs by augmenting the peak network load with a high frequency of ISS events that will be initiated by the attack.

 \item{\textbf{Compromise UEs.}} The attacker compromises a set of UEs and use them as botnet to launch the RSA/TSA. The strategic use of UEs as attack vectors exploits the inherent trust and legitimacy of these devices within the network, effectively masking the malicious intent and making  the attack detection more difficult.
 
 \item{\textbf{Launch RSA/TSA.}} To launch a RSA, the attacker identifies multiple random NSs (i.e., one per each compromised UE) and switches the compromised UEs to these NSs by triggering ISS for each of them. In contrast, to launch a TSA, the attacker switches the compromised UEs to a target, pre-selected NS.

\item{\textbf{Switch back compromised UEs to random NSs.}} The attacker switches back some or all the compromised UEs to random NSs in order to introduce random ISS patterns for UEs and make the attack look stealthy.

  \item{\textbf{Repeat steps (3) and (4).}} For increased and varying impact on the network and its NSs, the attacker can repeat steps (3) and (4) using the same or different subset of compromised UEs at different times, making the attack harder to detect while always causing disruptions to the network.
\end{enumerate}

\vspace{-0.3cm}

\vspace{-0.3cm}
\section{Inter-Slice Defender} \label{sec:InterSliceDefender}
%\vspace{-0.1cm}
%\begin{figure*}[h]
%    \centering
%    \includegraphics[width=0.7\linewidth]{images/modelInterSD_v2.png}
%    \captionsetup{justification=justified, singlelinecheck=false} 
%    \caption{\textcolor{blue}{Architecture of Inter-Slice Defender with data collection, feature extraction, and anomaly detection modules. It uses an LSTM-Autoencoder trained on benign data to detect attacks via reconstruction error and a threshold.}}
%    \label{fig:ApproachInterSD}
%\end{figure*}

\begin{figure*}[h]
    \centering
     \begin{subfigure}[h]{0.48\textwidth}
         \centering
         \includegraphics[width=\textwidth]{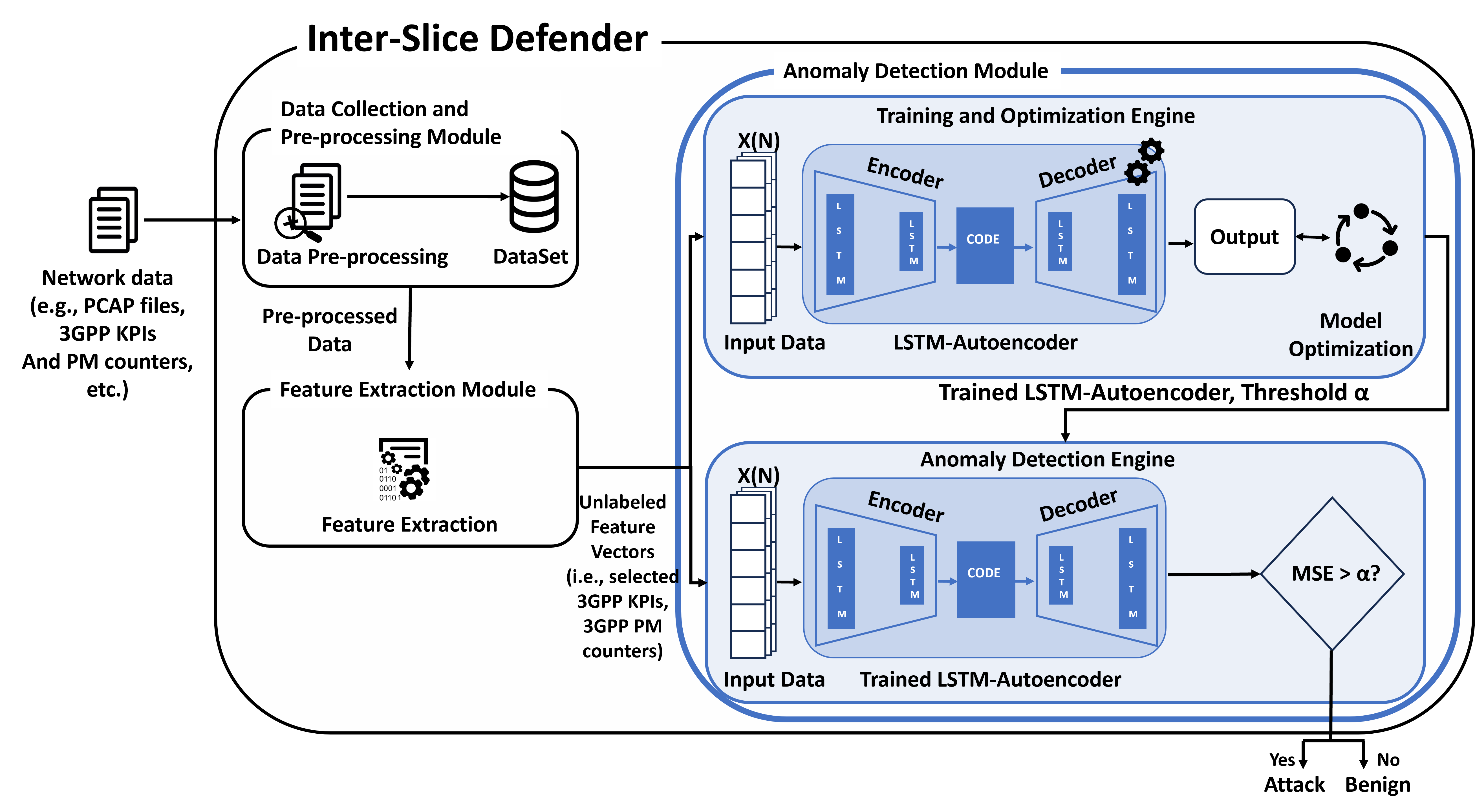}
         \caption{Inter-Slice Defender}
         \label{fig:inter_SD}
     \end{subfigure}
     \hfill
     \begin{subfigure}[h]{0.48\textwidth}
         \centering
         \includegraphics[width=\textwidth]{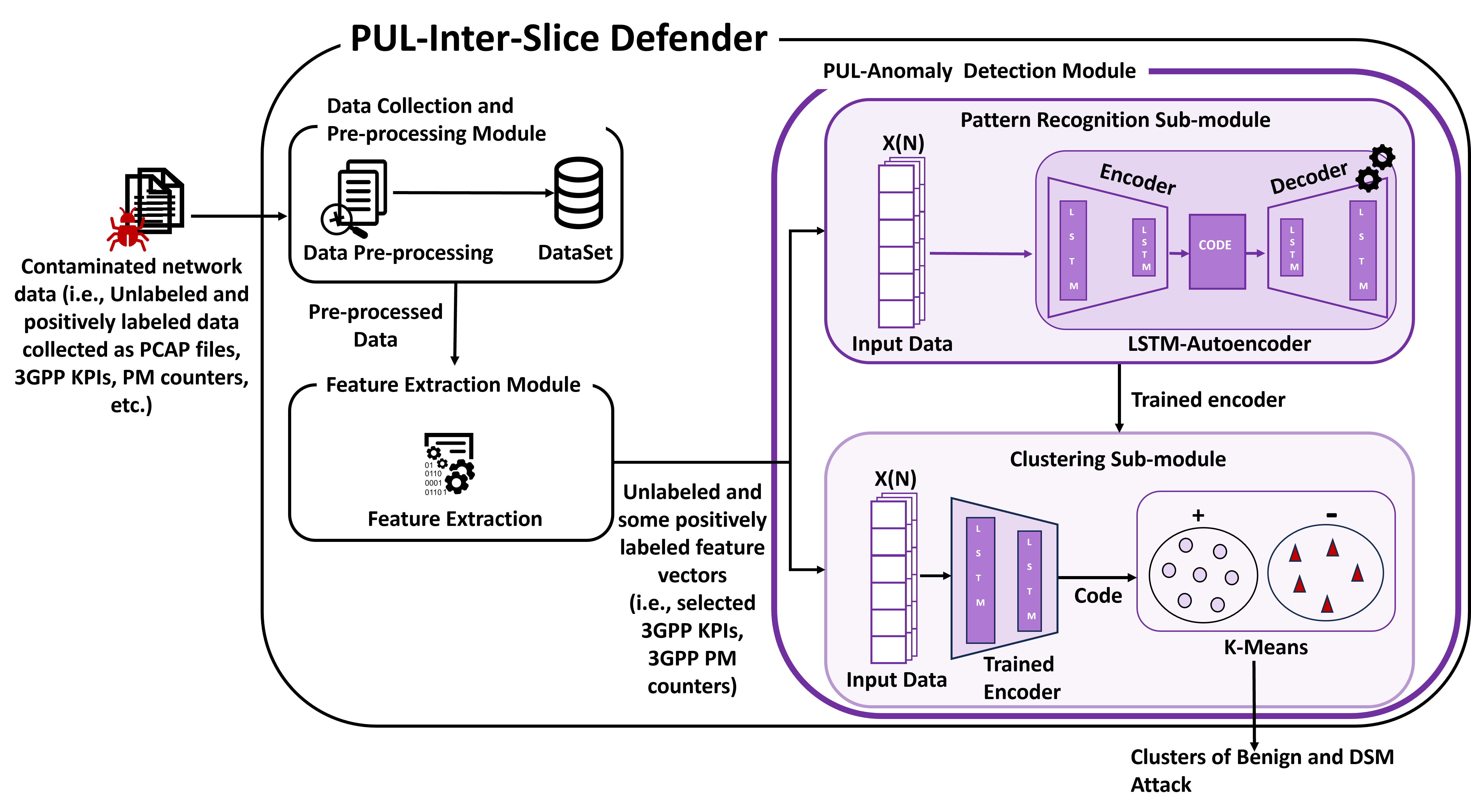}
         \caption{PUL-Inter-slice Defender}
         \label{fig:PUL_inter_SD}
     \end{subfigure}
     \vspace{0.5cm}
     \captionsetup{justification=justified, singlelinecheck=false} 
    \caption{\textcolor{black}{Architecture of Inter-Slice Defender and PUL-Inter-Slice Defender. Inter-Slice Defender uses an LSTM-Autoencoder trained on benign data to detect attacks via reconstruction error and a threshold, while PUL-Inter-Slice Defender adopts PUL by leveraging the latent space for feature extraction to classify benign and DSM attack samples in the presence of contaminated training data through K-means.} %\\\red{[HA: I think the difference between thr two figures is not clear visually. I suggest we give a color for each solution, and use for the different parts and the common parts keep them the same color.Also for PUL, you need to show that the trained encoder is sent rather than the latent space because you are getting the latent space from the encode in the clustering sub-module.]}} %is an enhanced version that leverages the latent space for feature extraction and employs a downstream classification task to detect RSA and TSA in the presence of contaminated training data.} 
    }
    \label{fig:AMFImpact}
\end{figure*}

%As this study extends the research conducted in our previous work \cite{rayalam2024}, this section provides a summary of its proposed solution, named Inter-Slice Defender, to detect DSM attacks, with a particular focus on the findings and the gaps identified in that study.

%Inter-Slice Defender (Figure \ref{fig:inter_SD}) is an NS anomaly detection solution that leverages an LSTM-Autoencoder to detect DSM attacks and their variations. The Inter-Slice Defender is composed of three modules: data collection and pre-processing module, feature extraction module, and anomaly detection module.

\textcolor{black}{ As this study extends the research conducted in our previous work \cite{rayalam2024}, this section presents its proposed solution , named Inter-Slice Defender (Fig. \ref{fig:inter_SD}), for detecting DSM attacks, with emphasis on its key findings and limitations. Fig. \ref{fig:inter_SD} illustrates this approach, which comprises three modules: data collection and pre-processing, feature extraction, and anomaly detection.}

The data collection and pre-processing module gathers 5G network data and prepares it for the feature extraction module, which extracts, selects, and normalizes 3GPP KPIs  \cite{3GPPKPI} and PM counters \cite{3GPP_PM} that are shown in Table \ref{tab:PM}.
The anomaly detection module is composed of two engines: the training and optimization engine, and the anomaly detection engine. The training and optimization engine is responsible for training the anomaly detection model based on an LSTM-Autoencoder \cite{said2020network}, optimizing its architecture and hyperparameters, and selecting a threshold that ensures good detection performance. The choice of LSTM-Autoencoder is based on LSTM's suitability for processing and learning from sequential data that capture long-term dependencies, while the Autoencoder is chosen for its ability to compress input data into a lower-dimensional latent space and then reconstruct it.
%Inter-Slice Defender leverages the fact that an Autoencoder is an unsupervised model that learns to reconstruct input data  \cite{said2020network}. When used for anomaly detection, it is trained primarily on data assumed as benign in majority and, as a result, fails to accurately reconstruct anomalous input. This failure leads to a significant difference between the input and its reconstructed version, known as the reconstruction error, which is used to identify anomalies by comparing it against a threshold. The anomaly detection engine uses the trained LSTM-Autoencoder model provided by the training and optimization engine, along with a selected threshold $\alpha$, to perform real-time anomaly detection. This detection is based on comparing the reconstruction error from the LSTM-Autoencoder against the threshold. % $\alpha$.
\textcolor{black}{Inter-Slice Defender leverages the Autoencoder’s ability as an unsupervised model to reconstruct input data \cite{said2020network}. Trained primarily on benign data, the model poorly reconstructs anomalous inputs, resulting in a high reconstruction error. This error, compared against a selected threshold  $\alpha$, is used by the anomaly detection engine to perform real-time anomaly detection.}

\textcolor{black}{Inter-Slice Defender (Fig. \ref{fig:inter_SD}) achieved notable results, reaching an average F1-score of 98.75\%, demonstrating its robustness in detecting DSM attacks and their variations \cite{rayalam2024}. However, this performance was based on training exclusively with benign data under the assumption of a clean training environment, which is unrealistic in real-world deployments where operational data may include noise or stealthy attacks. To better reflect practical scenarios, we retrained Inter-Slice Defender using contaminated datasets composed of typical operational data mixed with defined percentages of attack samples. This resulted in a significant performance drop, with the F1-score declining to 84.6\% when only 5\% of the training data was contaminated \cite{rayalam2024}. The presence of such anomalous data introduces conflicting patterns that compromise the model’s ability to learn a reliable representation of normal behavior. This weakens the discriminative power of the reconstruction error, as the model begins to internalize features of both benign and attack inputs. Consequently, the error gap narrows, leading to false negatives and underscoring the need for further adjustments to ensure robust anomaly detection in real-world network environments.}
\vspace{-0.3cm}

\section{PUL-Inter-Slice Defender}
\label{sec:methodology}

We present the PUL-Inter-Slice Defender (Fig. \ref{fig:PUL_inter_SD}), our novel NS anomaly detection solution that leverages LSTM-Autoencoder, K-Means, 3GPP KPIs \cite{3GPPKPI}, and PM counters \cite{3GPP_PM} (Table \ref{tab:PM}) to detect DSM attacks and their variations using a training dataset composed of positive and negative samples. Our approach relies on several key assumptions inherent to PUL, which are fundamental in handling this type of dataset. PUL-Inter-Slice Defender is composed of three modules: data collection and pre-processing module, feature extraction module, and PUL-anomaly detection module. We begin with a description of the PUL assumptions, followed by detailed explanations of each module.

\vspace{-0.5cm}
\subsection{PUL Assumptions}
\label{sec:PUassumptions}
The PUL is a binary classification problem where only a subset of labeled positive and unlabeled samples are available. This approach addresses the practical limitations of acquiring fully labeled datasets due to factors such as time, expertise, and financial investment \cite{papivc2023conditional,bekker2020learning}. The aim of PUL is to construct a classifier that can effectively distinguish between positive and negative samples, even when the training data consists of only a subset of labeled positive examples and a large set of unlabeled examples, thereby alleviating the cost of fully labeling data. It is important to note that the labeling process in PUL requires satisfying specific assumptions, as outlined in \cite{bekker2020learning}. In light of this, our approach incorporates several of these assumptions, which are enumerated subsequently:

\begin{enumerate}

    \item {\textbf{Separability:}}
The data used for PUL is assumed to be separable, that is positive and negative classes are naturally separable, indicating the existence of a classifier that can perfectly distinguish between the two. This assumption facilitates the development of approaches that concentrate on delineating a clear boundary between these two classes.

\item {\textbf{Smoothness:}}
This assumption states that samples close to each other in the feature space are likely to share the same label, thereby enabling the application of techniques such as clustering. This enhances learning by assuming that proximity in spatial or feature dimensions is indicative of label similarity.

    \item {\textbf{Unlabeled data contains positive and negative samples:}}
In PUL, only a subset of positive samples is labeled, while negative samples remain without explicit labels. Consequently, our unlabeled dataset comprises both positive samples that were not chosen for labeling and negative samples. This constitutes a core principle of PUL, differentiating it from other learning methods such as supervised learning, where all instances are explicitly labeled.

    \item {\textbf{Selected Completely At Random (SCAR):}}
This assumption implies that the labeled samples are a uniform subset of the positive samples, selected without bias towards any specific attributes, which simplifies the application of PUL by enabling it to be treated like a binary classification problem.
\end{enumerate}
%\vspace{-0.5cm}
\subsection{Data Collection and Pre-processing Module} \label{sec:DataCollection}
The data collection and pre-processing module (Fig. \ref{fig:PUL_inter_SD}) collects unlabeled data alongside a subset of positive labeled data (e.g., PCAP files, 3GPP KPI, PM counters, etc.) from the 5G network and pre-processes it to be ready for use by the feature extraction module. The collected data can include PCAP files depicting signaling messages between the 5G network components (i.e., Radio Access Network (RAN) and core NFs) and which pertain to the different 5G procedures such as ISS, registration, etc. It can also account for 3GPP KPIs and PM counters (Table \ref{tab:PM}) that are usually calculated at the different 5G NFs, and can be available at the Network Data and Analytics function (NWDAF) \cite{nwdaf}. NWDAF is a 5G NF that facilitates data collection and analysis. It can collect KPIs from the 5G NFs and calculate others such as those KPIs related to NSs \cite{nwdaf}.

The data collected by the data collection and pre-processing module then undergoes a transformation that simplifies the extraction of features. For instance, in this work, we use TShark \cite{Thark} to collect PCAP files from the network. We merge these files into a single one while maintaining their chronological order. This will serve the accurate calculation of time series based features by the feature extraction module. Then, we pre-process the unified PCAP file by extracting the most relevant information (e.g., IP source, IP destination, port destination, HTTP/2 header, etc.) for feature extraction.%We save them in a CSV file that is then shared with the feature extraction module.

%\begin{figure*}[h]
%    \centering
%    \includegraphics[width=0.7\linewidth]{images/model2_v2.png}
%    \captionsetup{justification=justified, singlelinecheck=false} 
%    \caption{\textcolor{blue}{PUL-Inter-Slice Defender: An enhanced version of Inter-Slice Defender that leverages the latent space for feature extraction and employs a downstream classification task to detect RSA and TSA in the presence of contaminated training data.}}
%    \label{fig:Approach}
%\end{figure*}
\begin{table*}[h]  % Optional: [h] places the table "here," adjust as needed
    \centering
    \caption{Features of PUL-Inter-Slice Defender model.}  % Add a caption to your table
   \resizebox{15.8cm}{!}{
    \begin{tabular}{|p{1.5cm}|p{6 cm}|p{8.9cm}|}  % Define the number of columns and their formatting
        
        \hline
         \textbf{Type}& \textbf{3GPP KPI features}& \textbf{Definition} \\  % Add column headers
%        \multicolumn{2}{|c|}{\textbf{3GPP KPI features}} & \textbf{Definition} \\ 
        \hline %\hline
        \multirow{4}{1.5cm}{3GPP-NS} & *Registration success rate of one single NS & Success ratio of registration procedures (i.e., ratio of number of successful registrations over total number of attempted registrations)  within a single NS for a specific AMF set.  \\  % Add data for row 1
        \cline{2-3}
        & *PDU session establishment success rate of one NS &  Rate of successful PDU session establishment request over total number of attempted requests across all SMFs associated with a specific NS. \\  % Add data for row 2
        \cline{2-3}
         & *Mean number of PDU sessions of network and NS& Average number of successful PDU session  within a specific NS.\\  % Add data for row 3
        \cline{2-3}
        & *Maximum number of PDU sessions of NS & Maximum number of successfully established PDU sessions within a single NS.   \\  % Add data for row 4        
        \hline \hline

       \multicolumn{1}{|c}{} & \multicolumn{1}{p{6cm}|}{\textbf{3GPP PM Counter features}} & \textbf{Definition} \\
       % \hline
       %& \textbf{3GPP PM Counter features} &  \textbf{Definition}\\  % Add column headers
%         \multicolumn{2}{|c|}{\textbf{3GPP PM Counter features}} & \textbf{Definition} \\ 
        \hline 
         \multirow{4}{1.5cm}{3GPP-AMF} & Number of initial registration requests & Total number of initial registration requests that AMF receives. \\  
          
        \cline{2-3}
         & Number of successful initial registrations & Count of successful initial registrations processed by the AMF.\\  % Add data for row 2
        \cline{2-3}
         &  Total number of attempted service requests & Number of attempted service requests including those initiated by the network and those initiated by UEs.\\  % Add data for row 3
        \cline{2-3}
         &  Total number of successful service requests & Cumulative count of successful service requests accounting for those initiated by both the network and by UEs.\\  % Add data for row 4
        \hline
         \multirow{5}{1.5cm}{3GPP-SMF} &  +Number of PDU session creation requests  & Number of PDU session creation requests received by the SMF.\\  % Add data for row 6
        \cline{2-3}
        & +Number of successful PDU session creations & Number of PDU sessions successfully established by the SMF.\\  % Add data for row 7
        \cline{2-3}
        & +Number of failed PDU session creations & Count of PDU sessions successfully created by the SMF.\\  % Add data for row 8
        \cline{2-3}
         &  *Max time of PDU session establishment & Maximum time for PDU session establishment in each granularity period divided into sub-counters for each NS.\\  % Add data for row 9        
    
       \cline{2-3}
        &   Number of released PDU sessions (AMF initiated) & Number of PDU sessions released at SMF with initiation originating from the AMF.\\  % Add data for row 9
        \hline
        \multirow{3}{1.5cm}{3GPP-NSSF} &   Number of NS selection requests & Total number of NS selection requests that the NSSF receives.\\  % Add data for row 9     
        \cline{2-3}
        &   Number of successful NS selections & Total successful NS selections executed by the NSSF.\\  % Add data for row 9
        \cline{2-3}
        &   Number of failed NS selections & Number NS selection attempts that failed at the NSSF.\\  % Add data for row 9
         \hline\hline
       \multicolumn{1}{|c}{} & \multicolumn{1}{p{6cm}|}{\textbf{Non-3GPP PM Counter feature}} & \textbf{Definition} \\
        \hline

%          & \textbf{Non-3GPP PM Counter feature}& \textbf{Definition} \\  % Add column headers
%        \multicolumn{2}{|c|}{\textbf{Non-3GPP PM Counter feature}} & \textbf{Definition} \\ 
         %\hline\hline
         AMF &  Number of failed initial registrations & Number of unsuccessful registrations. \\  % Add data for row 5
       \hline
    \end{tabular}
 }
 %\vspace{0.1cm}
    \label{tab:PM}  % Add a label for cross-referencing

    * Feature computed per each NS;   + Feature computed per each SMF
%\vspace{-0.5cm}
\end{table*}

\vspace{-0.5cm}
\subsection{Feature Extraction Module}%\vspace{-4pt}
\label{sec:fextrac} 
The feature engineering module (Fig. \ref{fig:PUL_inter_SD}) acquires the pre-processed data from the data collection and pre-processing module and uses it for feature extraction. This data is used for feature extraction, selection, and normalization to extract 3GPP KPI \cite{3GPPKPI} and PM counters \cite{3GPP_PM} such as those listed in Table \ref{tab:PM}. These features capture the 5G network and its NSs normal behavior, along with the abnormalities that can be caused by a DSM attack. The calculated features are time-series based, suitable for LSTM-Autoencoder required input format. To select the most pertinent features, a variance threshold process %\cite{sklearnVarianceThreshold}
is employed which eliminates features characterized by minimal variations or perceived as noise.

Consequently, a total of 17 distinctive features (Table \ref{tab:PM}) are selected to comprehensively scrutinize, analyze, and establish the normal and attack behavioral patterns exhibited by 5G networks. They can be classified per type that reflects their calculation. For instance, some features are calculated per NF (i.e., AMF, SMF, NSSF), mainly those NF involved in ISS related procedures (e.g., \enquote{Number of PDU session creation requests} feature is calculated at the SMF as it is involved in PDU session establishment procedure) while others are calculated per NS. With the exception of \enquote{Number of failed initial registrations} feature that is not standardized by 3GPP, the computation of the remaining features follows 3GPP specifications. These features can be collected from the NWDAF or the NFs if available or can be calculated from the PCAPs as in this work. Finally, it is worth noting that in the case where any of the aforementioned NFs is dedicated to an NS, its related features will then reflect the NS it serves.
The total number of features processed by this module contains information that represents the 5G network in normal conditions and under DSM attacks.

%\textcolor{red}{PUL-anomaly detection module} leverages our approach's training process, incorporating PUL. This module utilizes an LSTM-autoencoder architecture combined with K-means clustering, guided by a PUL strategy known as the two-step technique, which is used to label unlabeled samples. This technique, based on the assumptions of separability and smoothness \cite{bekker2020learning} \textcolor{red}{(section \ref{sec:PUassumptions})} , comprises two stages: \textcolor{red}{(1) extraction of negative, and optionally positive, samples from the unlabeled dataset, and (2) model training using these identified negatives alongside labeled positive samples.} These two steps are realized in conjunction with the \textcolor{red}{pattern recognition sub-module and the clustering sub-module of the PUL-anomaly detection module}, detailed hereafter.
%\vspace{-0.4cm}
\subsection{PUL-Anomaly  Detection Module
} \label{sec:ADM} 
The PUL-anomaly detection module (Fig. \ref{fig:PUL_inter_SD}) leverages our approach’s training process, incorporating PUL. This module utilizes an LSTM-autoencoder architecture combined with K-means clustering, guided by a PUL strategy that can be considered a variation of the two-step PUL technique. Standard two-step PUL techniques comprise two steps \cite{bekker2020learning}. The first step involves the extraction of negative, and optionally positive samples from the unlabeled dataset. The second step involves training a model using the identified negatives alongside labeled positive samples \cite{bekker2020learning}. However, in our approach, we first use the encoder of the LSTM-autoencoder to extract features (step 1) that characterize positive and negative samples, and then apply K-means clustering to separate these features into positive and negative clusters (step 2). These two steps are realized in conjunction with the pattern recognition sub-module and the clustering sub-module of the PUL-anomaly detection module, detailed hereafter.

\textbf{Pattern Recognition Sub-module.}
Pattern recognition sub-module
is developed to extract the most critical patterns, correlations, temporal dependencies, and informative features from those provided by the feature extraction module. It employs an LSTM-Autoencoder \cite{said2020network} to this end, training and preparing its encoder for effective feature extraction in the subsequent sub-module (i.e., the clustering sub-module). The selection of an LSTM-Autoencoder is based on its capability to effectively manage sequences and time-based patterns through LSTM. This capability is coherently integrated with an Autoencoder, which learns an abstract representation of its inputs (i.e., the normal behavior of the network and attack scenarios), accomplishing this through the compression and reconstruction of the input data \cite{said2020network}. The encoder effectively reduces the input dimensionality, thereby generating a latent space representation that captures the most significant characteristics and relationships in the training dataset, while the decoder is tasked with reconstructing the original input data from the compressed representation generated by the encoder. Thus, LSTM-Autoencoder yields a good depiction of the 5G network under both normal conditions and DSM attack scenarios, where the dependencies and precedence constraints of the various 5G procedures that represent ISS events are thoroughly detailed. For instance, an ISS event to a new NS cannot occur before deregistering from the current NS and registering with the new one (Section \ref{sec:dataSimulation}). Furthermore, the frequency of normal network behavior and attack patterns over time can be captured by an LSTM through its cell capabilities, which can remember values over various time intervals \cite{said2020network}.

\textbf{Clustering sub-module.}
%The clustering sub-module is designed to identify and classify unlabeled data as positive or negative \textcolor{red}{samples to detect DSM attacks.}
The clustering sub-module is designed to identify and classify unlabeled data as positive or negative samples, alleviating the problem of learning the decision boundary without a fully labeled dataset and achieving the goal of the PUL approach. Additionally, this module is employed to detect DSM attacks.

Upon completing the training phase of the LSTM-autoencoder model in the pattern recognition sub-module, its encoder component is employed by the clustering sub-module to extract features from the data processed by the feature extraction module, which characterized the unlabeled data and the available subset of positively labeled data. Inside the encoder, these features are compressed and encoded into a compact and meaningful representation that preserves the essential characteristics of the sequential data that represent normal and attack behavior. This refined representation %, extracted from the latent space that holds the compressed vector representations (feature vectors),
is leveraged by the K-means clustering algorithm. K-means clustering is an unsupervised machine learning algorithm, simple and easy to implement, and widely used in anomaly detection applications \cite{jain2022k}. It is designed to partition a dataset into K distinct, non-overlapping clusters, assigning each data point to one of the K clusters based on predefined similarity measures. By iteratively updating cluster centroids and reallocating data points, K-means effectively identifies the underlying structure within the data.

In our approach, K-means clustering algorithm  leverages the extracted features from the encoder to enhance its performance by improving clustering accuracy, thereby potentially augmenting the separability of the samples, which aligns with the smoothness and separability assumptions described in Section \ref{sec:PUassumptions}. This process facilitates the grouping of the data into two distinct clusters: one comprising positive samples (i.e., benign instances), and the other encompassing negative samples (i.e., attack instances).
%Furthermore, given access to a subset of positively labeled samples according to the SCAR assumption (Section \ref{sec:PUassumptions}), the clusters are subsequently mapped to the actual labels by identifying where the positive labeled samples ended up in terms of clustering. After that, each sample is labeled based on its cluster membership.
%\textcolor{red}{Furthermore, given the inclusion of the initial subset of positive labeled data in the training dataset, which is labeled according to the SCAR assumption (Section \ref{sec:PUassumptions}), this subset is used to determine which clusters has to be classified as positive and negative.}
Furthermore, given the inclusion of an initial subset of positively labeled data within the training dataset (Section \ref{sec:dataset}), it becomes feasible to assign labels to each cluster. By mapping these labeled samples to the resulting clusters, this subset facilitates the determination of which cluster has to be designated as positive and which as negative.

Once the K-means training is completed, new data entries are processed through the predefined sequence involving LSTM-Autoencoder’s encoder for feature extraction followed by clustering. The classification of a new data point, whether it is anomalous or not, is then assessed based on its association with a specific cluster. This assessment relies on the premise that the clusters distinctly represent either normal or anomalous instances.

\vspace{-0.3cm}
\section{5G Environmental Setup And Data Generation}
\label{sec:environmentalSetup}
Given the lack of a 5G dataset suitable for training and testing our PUL-Inter-Slice Defender solution, we present, in this section, our environmental setup and data generation strategy. We highlight our 5G testbed that we deploy to emulate normal network traffic in addition to RSA and TSA. 
\vspace{-0.5cm}
\begin{figure}[h]
    \centering
    \includegraphics[width=1\linewidth]{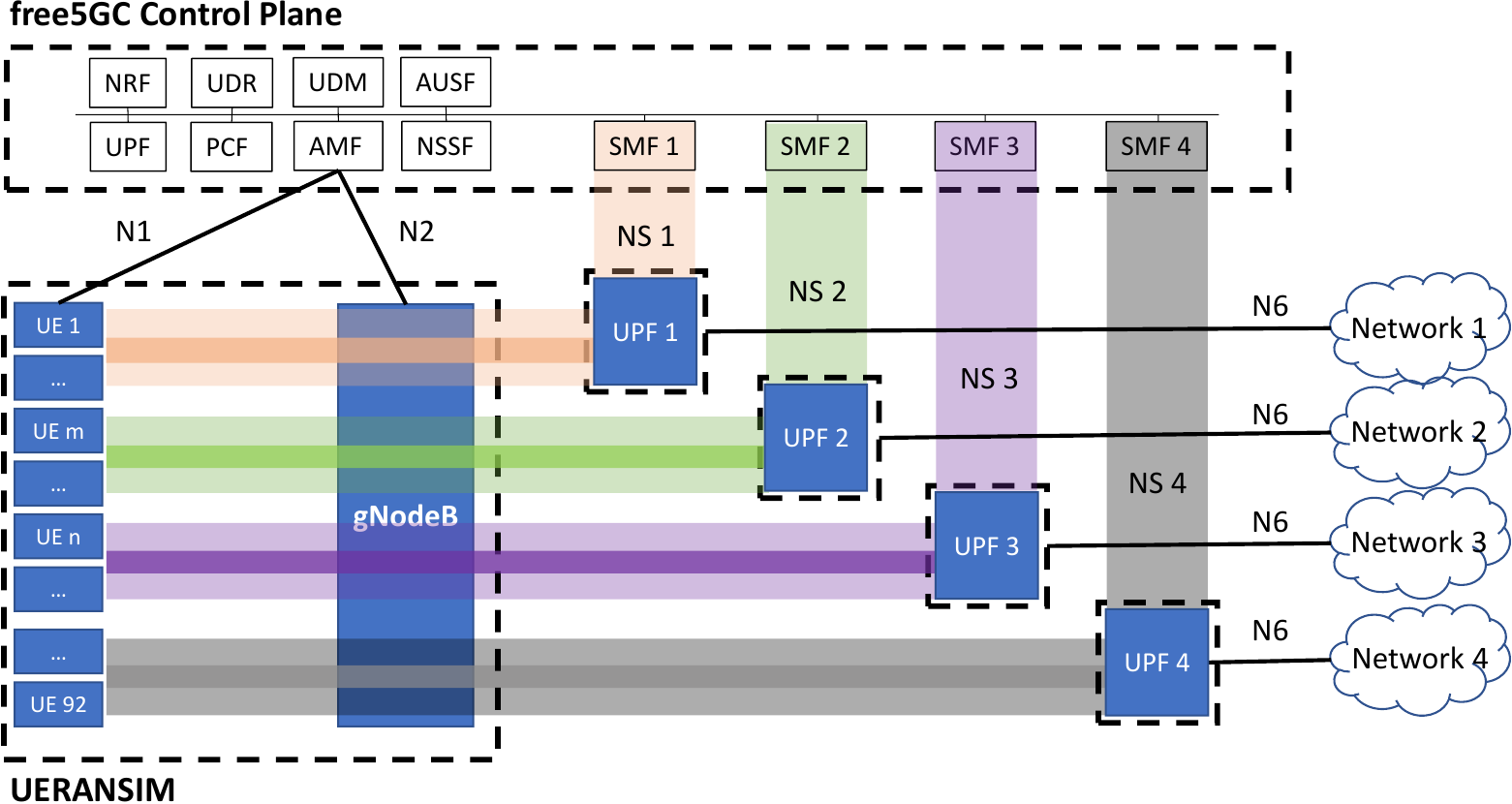}
    \captionsetup{justification=justified, singlelinecheck=false}
    \caption{5G testbed \textcolor{black}{comprising four NSs, designed in accordance with the 3GPP standard.}}
    \label{fig:Testbedmodel}
\end{figure}

\subsection{5G Testbed} \label{sec:testbed}

We employ the open source free5GC-compose and free5GC all-in-one implementation \cite{free5GC}, adhering to the 3GPP standard \cite{3GPP} to build our testbed, depicted in Fig. \ref{fig:Testbedmodel}.
The testbed runs free5GC-compose version 3.3.0 as the CP on a Virtual Machine (VM)  operating Ubuntu 20.04 – Focal. The VM features 8 virtual CPUs, 8 GB of RAM, and a 60 GB of hard drive, with each NF encapsulated within this virtualized environment. 
To enhance the realism of our emulation environment, we opt to segregate the RAN from the CP. This segregation involves installing UERAMSIN 3.2.6 \cite{UERANSIM_Aligungr}, a UE and RAN simulator, on a distinct VM. %As illustrated in Figure \ref{fig:testbed}[X],%
Our designed testbed comprises four NSs, each featuring a dedicated UPF installed on a separate VM, and a dedicated SMF. For the installation and configuration of all UPFs, we use the free5GC all-in-one version 3.3.0 \cite{free5GC} while the SMFs are deployed on containers in the VM hosting the CP. Furthermore, our testbed hosts 92 UEs configured to be able to connect to the existing four different NSs. The creation and management of the VMs within our solution are orchestrated through OpenStack \cite{OpenStack_Docs}.
 \vspace{0.5cm}
\begin{table}[h]  % Optional: [h] places the table "here," adjust as needed
    \centering
    \caption{Logical dependency between 5G procedures.}  % Add a caption to your table
   \resizebox{8.8cm}{!}{
    \begin{tabular}{|c|c|}  % Define the number of columns and their formatting
        \hline
        \textbf{Triggered procedure} & \textbf{Possible subsequent procedures}  \\  % Add column headers
        \hline
        ISS  & ISS, Uplink, Downlink, UE release PDU session,  gNodeB release PDU session  \\  % Add data for row 1
        \hline
        Registration & ISS, Uplink, Downlink, UE release PDU session,  gNodeB release PDU session  \\  % Add data for row 2
        \hline
        Uplink & ISS, Downlink, UE release PDU session,  gNodeB release PDU session \\  % Add data for row 3
        \hline
        Downlink  & ISS, Uplink, UE release PDU session,  gNodeB release PDU session  \\  % Add data for row 4
        \hline
        UE release PDU session  & ISS, Downlink, Uplink,  gNodeB release PDU session  \\  % Add data for row 5
        \hline
         gNodeB release PDU session  & ISS, Uplink, Downlink  \\  % Add data for row 6
%        \hline
%        Deregistration & ------ \\  % Add data for row 7
        \hline
    \end{tabular}
} 
    \label{tab:mytable}  % Add a label for cross-referencing
\end{table}
 \vspace{-0.5cm}

\begin{figure*}[t]
    \centering
    \begin{subfigure}[b]{0.23\textwidth}  % Adjust the width to fit all images horizontally
        \centering
        \includegraphics[width=\textwidth]{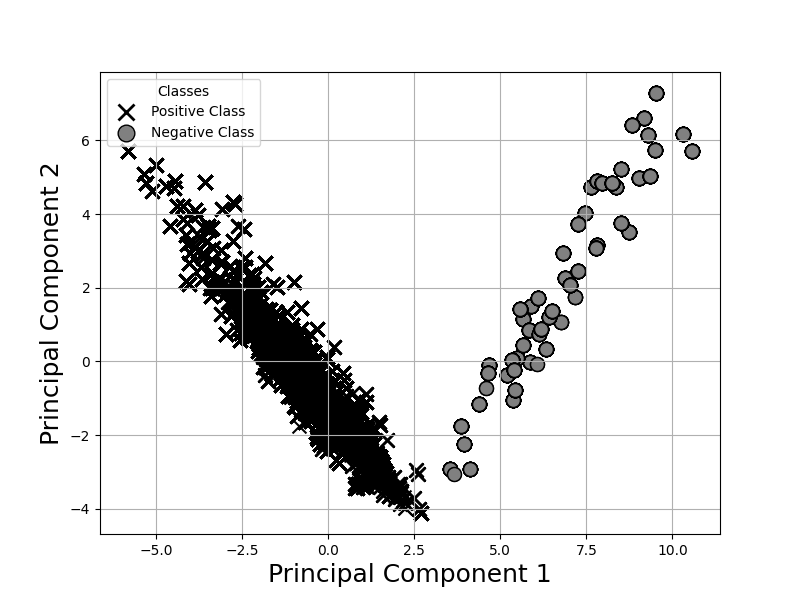}
        \caption{10\% of negative samples}
        \label{fig:pca10}
    \end{subfigure}
    \hfill
    \begin{subfigure}[b]{0.23\textwidth}
        \centering
        \includegraphics[width=\textwidth]{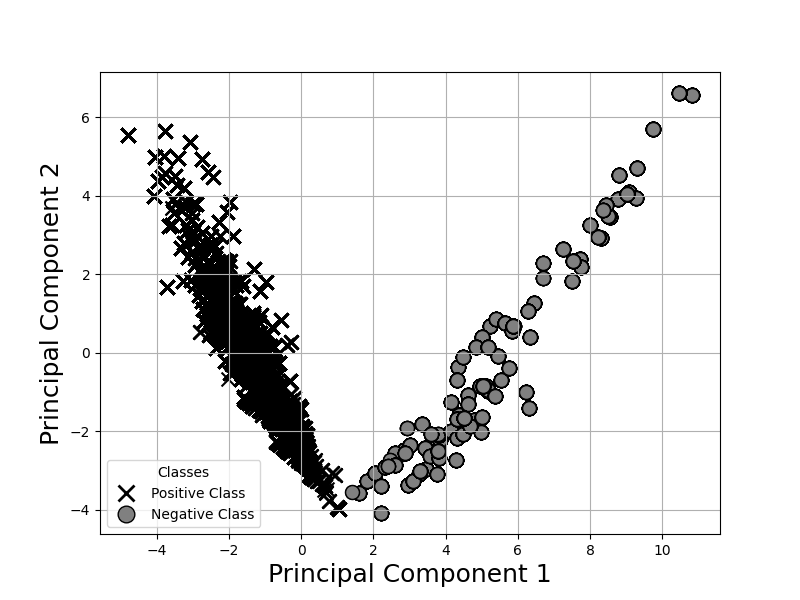}
        \caption{20\% of negative samples}
        \label{fig:pca20}
    \end{subfigure}
    \hfill
    \begin{subfigure}[b]{0.23\textwidth}
        \centering
        \includegraphics[width=\textwidth]{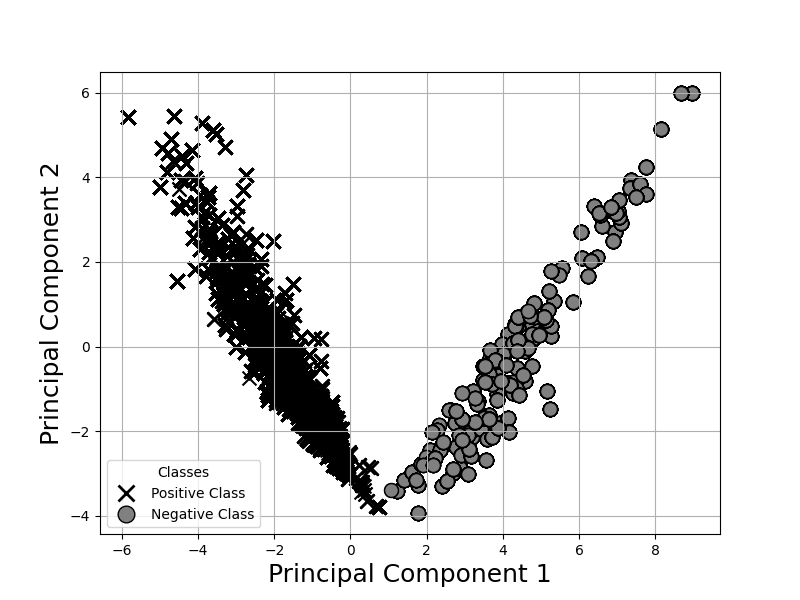}
        \caption{30\% of negative samples}
        \label{fig:pca30}
    \end{subfigure}
    \hfill
    \begin{subfigure}[b]{0.23\textwidth}
        \centering
        \includegraphics[width=\textwidth]{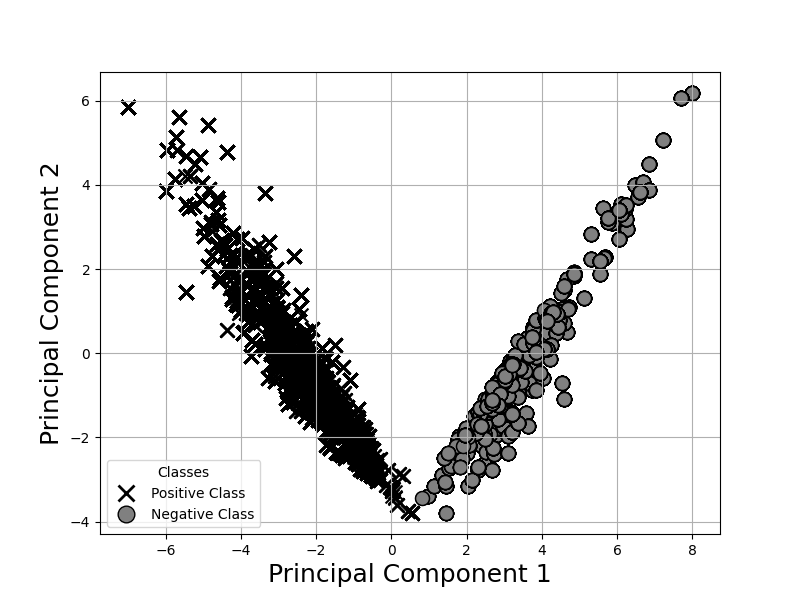}
        \caption{40\% of negative samples}
        \label{fig:pca40}
    \end{subfigure}
    \vspace{0.5cm}
    \caption{2D PCA plot showing positive and negative classes in the training dataset with different levels of negative samples}
    \label{fig:PCASoverall}
\end{figure*}
%\vspace{-0.3cm}
\subsection{Emulation of Normal and Attack Network Traffic}
\label{sec:dataSimulation}

To collect benign and attack data, we perform three different emulations. The initial emulation replicates a 5G network operating under normal conditions without any external attacks. In contrast, the subsequent two emulations were conducted to assess the impact of RSA and TSA.

\textbf {Normal network traffic emulation.} Throughout the 120-minute duration dedicated to this emulation, 92 UEs are used to emulate varying loads of normal network traffic. Each UE randomly triggers many 5G procedures from those detailed in Table \ref{tab:mytable}. Notably, the 5G procedures listed under the \enquote{Possible subsequent procedures} column are executable solely and exclusively if the 5G procedure indicated under the \enquote{Triggered procedure} column has been successfully completed. This logical interdependence underscores the sequential nature of these procedures. For instance, as outlined in Table \ref{tab:mytable}, following the successful completion of a registration procedure, several events can be triggered: ISS, Uplink, Downlink, UE release PDU session, and gNodeB release PDU session.
Finally, the benign activity of the UEs is recorded in PCAP files. The latter constitutes the benign dataset that will be used for our LSTM-Autoencoder model training and testing.
%\vspace{0.3cm}

%\vspace{-0.3cm}
\textbf {RSA and TSA emulations.}
To conduct the RSA and TSA emulations, we adhere to the same approach as in the benign emulation. We perform the emulation of each attack using a total of 92 connected UEs out of which 28 are compromised and used for the attacks. Following our threat model (Section \ref{sec:DSMattack}), the attacker strategically decides on the time to launch the RSA or TSA such that it coincides with the network’s peak activity. % to increase the impact of the executed attack.
As a result, during the attack emulations, the network operates normally for the first 60 minutes, after which the attack is initiated when the load on the network is designed to be at its peak. When the attacks start, the 28 compromised UEs are connected to any of the four NSs and are used to perform ISS events in the quest of overloading the CP.

\vspace{0.5cm}
\begin{table}[htb!]
\centering
\caption{Datasets statistics.}
\resizebox{8.8cm}{!}{
\begin{tabular}{|l|c|c|c|c|c|}
\hline
\textbf{Dataset type} & \textbf{Contamination} & \textbf{No. of records} & \textbf{Benign records} & \multicolumn{2}{c|}{\textbf{Attack records}} \\ \cline{5-6} 
                      &                        &                            &                         & \textbf{RSA}            & \textbf{TSA} \\ \hline
Training Dataset      & 10\%       & 40000                      & 36000                   & 2000                       & 2000           \\ \hline
Training Dataset      & 20\%       & 40000                      & 32000                   & 4000                       & 4000           \\ \hline
Training Dataset      &30\%       & 40000                      & 28000                   & 6000                       & 6000           \\ \hline
Training Dataset      & 40\%       & 40000                      & 24000                   & 8000                       & 8000           \\ \hline
RSA Test dataset      & ---           & 20000                      & 10000                   & 10000                   & 0           \\ \hline
TSA Test dataset      & ---           & 20000                      & 10000                   & 0                       & 10000       \\ \hline
%\textcolor{red}{RSA + TSA Test} & Contaminated & 20000 & 10000 & 5000 & 5000 \\ \hline
\end{tabular}
}
\label{tab:DataForTests}
%\textcolor{blue}{\\ * It will be contaminated with different percentages of attack samples.  \\ * A few benign samples are labeled}
\vspace{-0.5cm}
\end{table}

\subsection{Datasets for Anomaly Detection Model}\label{sec:dataset}

The data generated from the aforementioned emulations is used to create different datasets to facilitate the training and evaluation of PUL-Inter-slice Defender model.

\subsubsection{Training Dataset} 
To develop an effective anomaly detection solution for contaminated data, we employ four training datasets (Table \ref{tab:DataForTests}) encompassing different proportions of
both benign samples, representing the positive class, and DSM attack samples (i.e., RSA and TSA samples), representing the negative class. Specifically, the proportions of attack samples are set at 10\%, 20\%, 30\%, and 40\% of the total training dataset, with the remaining samples being benign. Each dataset includes 10\% labeled benign data, identified by a security expert who randomly selects a diverse and representative subset of positive samples, ensuring unbiased and high-quality data. This enhances the effectiveness of our PUL-anomaly detection module’s binary classifier, as established by the SCAR assumption (Section \ref{sec:PUassumptions}). It also aligns with the PUL assumption, which asserts that a dataset contains a subset of labeled positives and unlabeled samples, where the latter comprises both positive and negative samples (Section \ref{sec:PUassumptions}).

%identified by a security expert. The labeling process is carried out by a security expert who meticulously identifies a diverse and representative selection of positive samples from the training dataset. These samples are randomly selected to avoid biases related to specific data attributes, ensuring a high-quality and representative data subset that enhances the effectiveness of our PUL-anomaly detection module's binary classifier, as established by the SCAR assumption (Section \ref{sec:PUassumptions}).
%It also aligns with the PUL assumption, which asserts that a dataset contains a subset of labeled positives and unlabeled samples, where the latter comprises both positive and negative samples (Section \ref{sec:PUassumptions}).
%Consequently, this approach ensures that the resulting subset of data is both high-quality and representative, thus enhancing the effectiveness of our PUL-anomaly detection module in implementing a binary classifier, as established by the SCAR assumption (Section \ref{sec:PUassumptions}). Additionally, this approach also aligns with the PUL assumption, which asserts that a dataset contains only labeled positive and unlabeled samples, where the latter comprises both positive and negative samples (Section \ref{sec:PUassumptions}).

Furthermore, to analyze the distribution and assess the class separability and smoothness assumptions within our training dataset, we employed Principal Component Analysis (PCA) \cite{greenacre2022principal}. PCA reduces the dataset’s dimensionality to two dimensions, enabling the visualization of data in a scatter plot to verify the separability of classes in a reduced-dimensional space. Fig. \ref{fig:PCASoverall} illustrates a clear separation between positive and negative classes across all four training datasets, indicating that they are distinguishable within the transformed feature space. This suggests that training a model with positive and unlabeled data could effectively differentiate between positive and negative samples. 
Fig. \ref{fig:PCASoverall} also demonstrates that our training datasets meet the smoothness assumption across the different proportions of positive and negative samples. Both classes, positive and negative, exhibit compact clustering within each class, indicating similarity and supporting this principle. Furthermore, the boundary between classes is fairly distinct with minimal overlap, reinforcing that data points of the same class are closer to each other than to those of the other class.

\subsubsection{Test datasets}
To rigorously evaluate our anomaly detection model, we use two test datasets, each containing benign and malicious data from RSA and TSA attack emulations (Section \ref{sec:dataSimulation}). This approach thoroughly assesses the model's ability to distinguish between normal and anomalous patterns.

It is crucial to highlight that the training and test datasets are meticulously designed to be mutually exclusive, ensuring that there are no redundant records between them.
 \vspace{-0.3cm}

\section{DSM Attack Impact on 5G Control Plane }\label{sec:DSM_onCP}
To assess the impact of the RSA and TSA on the 5G network, we observe the CPU utilization of the different 5G CP NFs during RSA and TSA emulations and compare it with their CPU utilization during normal network traffic. 
\begin{figure*}[h]
    \centering
     \begin{subfigure}[h]{0.32\textwidth}
         \centering
         \includegraphics[width=\textwidth]{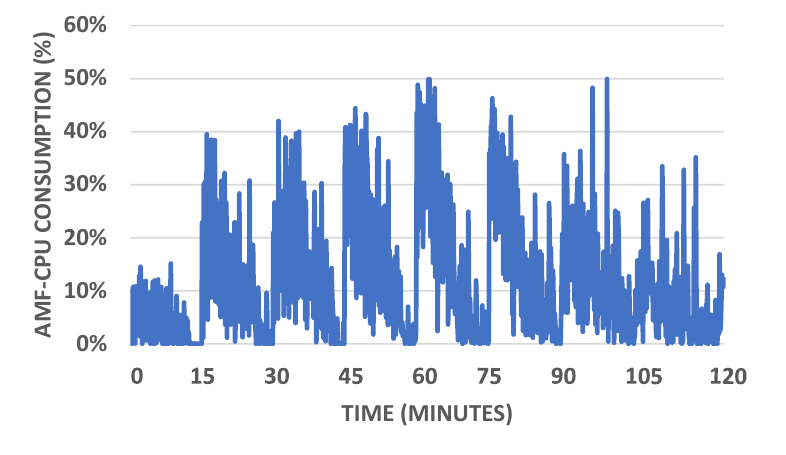}
         \caption{AMF - Benign emulation}
         \label{fig:amfBenign}
     \end{subfigure}
     \hfill
     \begin{subfigure}[h]{0.32\textwidth}
         \centering
         \includegraphics[width=\textwidth]{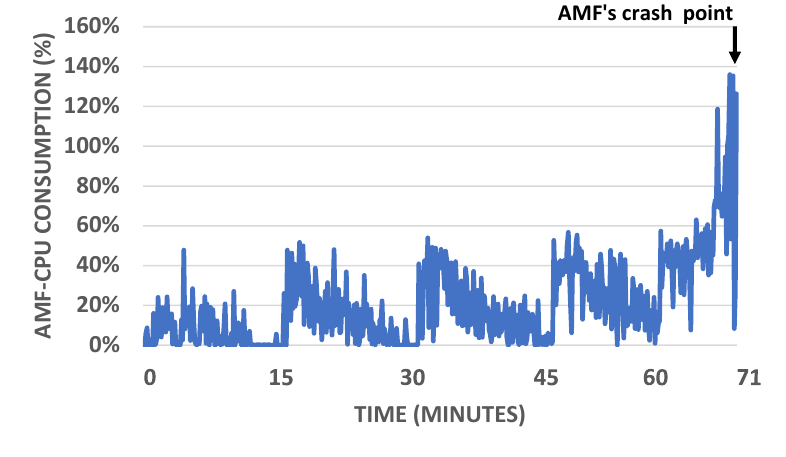}
         \caption{AMF - RSA}
         \label{fig:amfrsa}
     \end{subfigure}
   \hfill
 \begin{subfigure}[h]{0.32\textwidth}
         \centering
         \includegraphics[width=\textwidth]{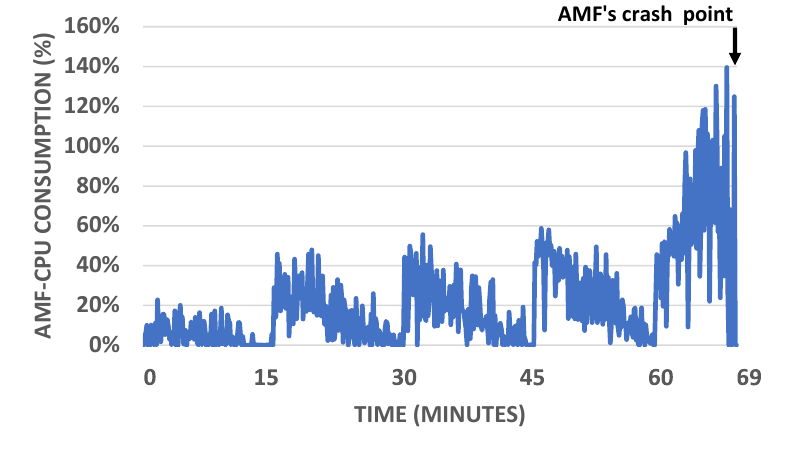}
         \caption{AMF - TSA}
         \label{fig:amftsa}
     \end{subfigure}     
    \vspace{0.5cm}
    \caption{AMF CPU consumption during benign and attack emulations.}
    \label{fig:AMFImpact}
\end{figure*}
%%%%%%%%%%%%%%%%%%%%
%\vspace{-0.3cm}
\subsection{Impact on AMF}\label{subsec:AMFImpact}

We first focus on the impact of the attacks on the AMF given that it is the CP NF that is involved the most in UE to 5G network communication and is the first CP point of contact. In fact, despite the role AMF plays in UE registration, authentication, and NS selection and allocation, it is usually shared among different NSs. Thus, when a significant number of UEs simultaneously perform ISS, such as in the case of RSA and TSA, we observe a significant increase in the AMF CPU utilization which leads to a DDoS (Fig. \ref{fig:AMFImpact}). 

As explained in Section \ref{sec:dataSimulation}, we emulate varying loads of normal network traffic and trigger the RSA and TSA at the peak load depicted at $time ~=~60~minutes$. Fig. \ref{fig:amfBenign} shows the AMF CPU utilization during normal network traffic emulation and depicts the highest utilization of $49.95\%$ at $time~=~60~minutes$. In contrast, Fig. \ref{fig:amfrsa} and \ref{fig:amftsa} show that the AMF CPU consumption reaches $135.98\%$ and $139.55\%$ during the RSA and TSA respectively which are launched at the peak network load (i.e., $time~=~60~ minutes$). The figures show that RSA and TSA last for $11~minutes$ and $9~minutes$ respectively before the crash of the AMF, and hence the whole network. This shows the DDoS impact the DSM attack can have on the 5G network and further emphasizes the need to secure 5G networks against it. Finally, note that AMF CPU utilization exceeded 100\% in our testbed due to containerized CP NFs sharing unused CPU resources. Without this sharing, the DDoS impact would have been observed earlier.

%Finally, it is worth noting that the AMF CPU utilization could exceed $100\%$ in our testbed because our CP NFs are containerized and can borrow from each other the CPU unused resources. Nonetheless, if the network does not allow sharing of unused CPU resources, the DDoS impact would have been observed earlier.
\vspace{-0.3cm}
\subsection{Overall Impact on 5G CP NFs}\label{subsec:NFImpact}

\begin{figure}[h]
    \centering
    \includegraphics[width=0.9\linewidth]{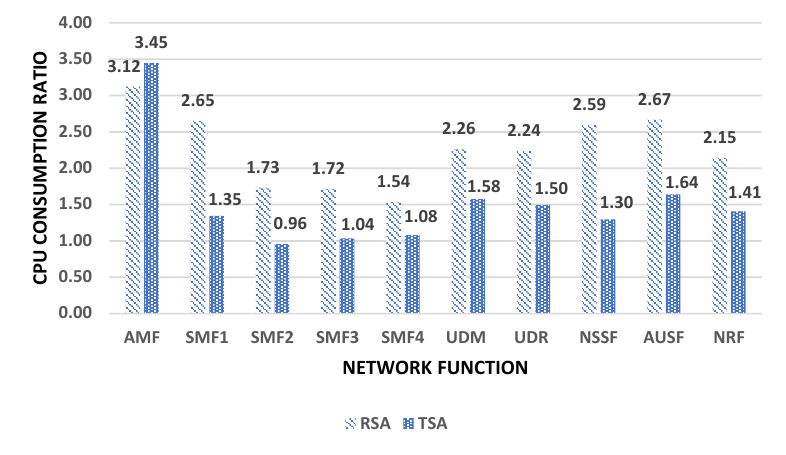}
    \captionsetup{belowskip=3pt}
    \caption{CPU consumption ratio during RSA and TSA.}
    \label{fig:cpuRatios}
   \vspace{-0.3cm}
\end{figure}

Despite the significant impact RSA and TSA have on the AMF, we observe a similar increase in the CPU consumption on other CP NFs as shown in Fig. \ref{fig:cpuRatios}. In fact, this figure shows the average CPU ratio for all the CP NFs during both RSA and TSA emulations. % (i.e., starting at $time~=~60~ minutes$ till the crash of the network compared to the benign simulation).
This ratio is calculated by dividing the average CPU consumption for each NF during the attack period (i.e., [$60~minutes - 71~minutes$] for RSA, [$60~minutes - 69~minutes$] for TSA) over that during the benign emulation for that same period.

Note that except for the AMF, the increase in the CPU utilization for all the NFs is greater during the RSA  than during the TSA. Fig. \ref{fig:cpuRatios} shows that RSA has a bigger impact on the 5G CP NFs. However, TSA has a higher impact on the AMF than the RSA, which results in degrading its performance in handling the requests, many of which ended up being dropped. Thus, fewer requests were forwarded to other CP NFs during TSA than during RSA which explains a smaller increase in their CPU utilization. Finally, unlike the AMF which CPU consumption exceeds $100\%$ during the attacks, we observe that CPU consumption of other CP NFs remains under $50\%$.

%\vspace{-0.7cm}
\section {Experimental Results} \label{sec:expResults}
%\begin{figure*}[h]
%    \centering
%    \begin{subfigure}[h]{0.45\textwidth}
%         \centering
%         \includegraphics[width=\textwidth]{images/RSA_TSA.png}
%         \caption{Inter-Slice Defender}
%         \label{fig:RSA_TSA}
%     \end{subfigure}
%     \hfill
%     \begin{subfigure}[h]{0.45\textwidth}
%         \centering
%         \includegraphics[width=\textwidth]{images/PUL_XGBoosT.png}
%         \caption{PUL-OCSVM/XGBoosT}
%         \label{fig:P_XGBoosT}
%     \end{subfigure}
%         \\ % Blank line
%          \vspace{0.7cm} 
%     \begin{subfigure}[h]{0.45\textwidth}
%         \centering
%         \includegraphics[width=\textwidth]{images/PUL_RF.png}
%         \caption{PUL-ocSVM/RF}
%         \label{fig:P_RF}
%     \end{subfigure}
%   \hfill
%    \begin{subfigure}[h]{0.45\textwidth}
%         \centering
%         \includegraphics[width=\textwidth]{images/PU_RSA_TSA.png}
%         \caption{PUL-Inter-Slice Defender}
%         \label{fig:PU_RSA_TSA}
%     \end{subfigure}
%    \vspace{0.5cm} % Add some vertical space between the subfigures
%    \captionsetup{justification=justified, singlelinecheck=false} 
%     \caption{\textcolor{blue}{Comparison of RSA and TSA detection performance under varying contamination levels across four models: (a) Inter-Slice Defender, (b) PULInter-Slice Defender, (c) PUL RF, and (d) PUL - XGBoost.}}
%    \label{fig:AMFImpactContamination}
%   \end{figure*}
\begin{figure*}[h]
    \centering
    \includegraphics[width=0.8\linewidth]{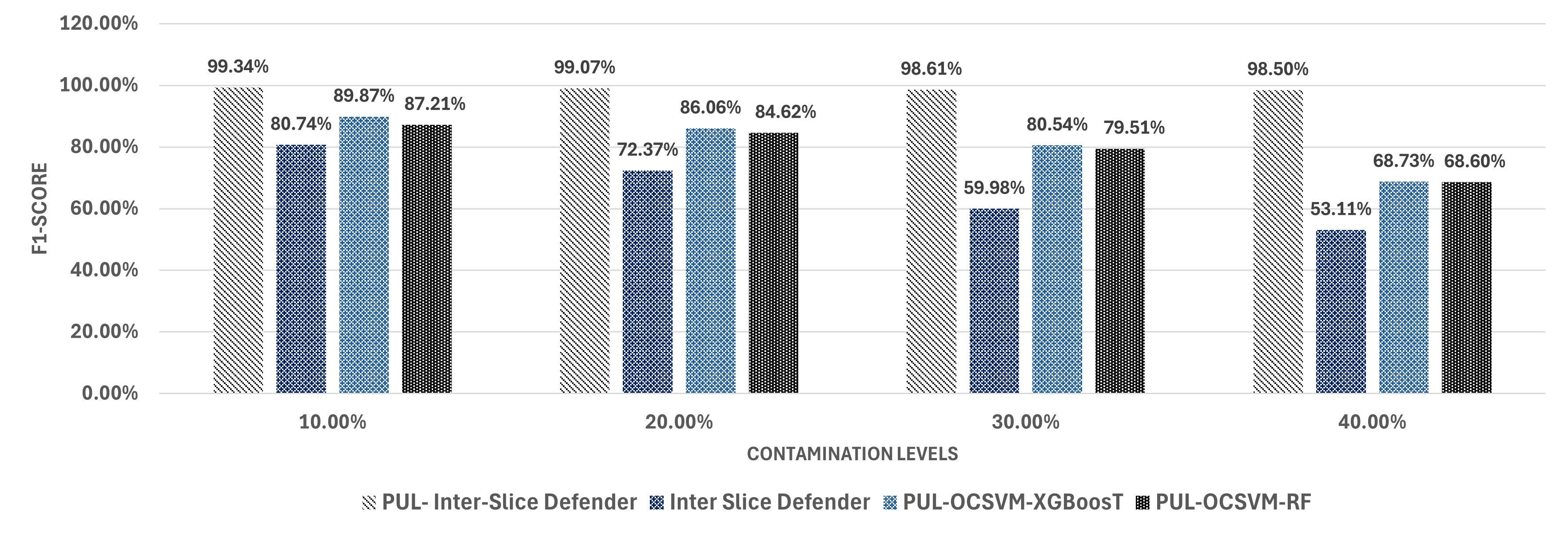}
    \captionsetup{justification=justified, singlelinecheck=false}
    \caption{\textcolor{black}{F1-score comparison of PUL-Inter-Slice Defender, Inter-Slice Defender, PUL-OCSVM-RF, and PUL-OCSVM-XGBoost for detecting RSA and TSA under varying training data contamination levels (10\%–40\%).} %\red{[HA: the figure is good but i feel the quality is not good. perhaps you need to save it again and add it? Also, just like the other figures, write the title for x and y axis in upper case.]}
    }  
    %\vspace{0.3cm}
    \label{fig:averages}
   % \vspace{-0.1cm}
\end{figure*}

In this section, we evaluate PUL-Inter-Slice Defender and assess its effectiveness in detecting RSA and TSA when its training dataset contains samples reflecting normal behavior in the 5G network and samples indicating DSM attacks.
\vspace{-0.3cm}
%\begin{table}[h]  % Optional: [h] places the table "here," adjust as needed
%    \centering
%    \caption{LSTM-Autoencoder hyperparameters.}  % Add a caption to your table
%    \begin{tabular}{|c|c|}  % Define the number of columns and their formatting
%        \hline
%        \textbf{Hyperparameter} & \textbf{Value}  \\  % Add column headers
%        \hline
%        Epochs  & 50 \\  % Add data for row 5
%        \hline
%        Dropout  & 0.2 \\  % Add data for row 6
%        \hline
%        L1 regularization & 0.01 \\  % Add data for row 7
%        \hline
%        Batch size & 32 \\  % Add data for row 7
%        \hline
%        Loss Function & MSE \\  % Add data for row 7
%        \hline
%        Learning rate & 0.01 \\  % Add data for row 7
%        \hline
%        Optimizer & Adam \\  % Add data for row 7
%        \hline
%        Hidden activation function & Relu \\  % Add data for row 7
%        \hline

%    \end{tabular}
    
%    \label{tab:Hyper}  % Add a label for cross-referencing
    %\vspace{-0.5cm}
%\end{table}

%\vspace{-0.5cm}
\subsection {LSTM-Autoencoder Architecture Selection} 
\label{sec:AnomalyDetectionSol}

To determine the optimal architecture for the PUL-Inter-Slice Defender’s LSTM-Autoencoder, which effectively learns patterns and dependencies in the sequential data representing normal and attack behaviors in our training dataset, we train and validate multiple architectures and evaluate their performance. To this end, we use a training dataset composed of 30\% attack data and 70\% benign samples (Table~\ref{tab:DataForTests}), as a dataset with a moderate proportion of negative samples allows our proposed approach to learn a more diverse representation of what constitutes a negative sample, which is crucial in PUL for avoiding false positives and preventing the model from overfitting to the positive class. Thus, we allocate 20\% of the training dataset as a validation dataset and train the model using its remaining portion (Table~\ref{tab:DataForTests}).
To select the model hyperparameters, we apply the K-fold cross-validation technique~\cite{CROSSv2}, which helps prevent overfitting and thoroughly evaluates model performance. \textcolor{black}{Specifically, we train the model using a batch size of 32, a learning rate of 0.01, and the Adam optimizer, with a dropout rate of 0.2 to mitigate overfitting. We use ReLU as the hidden activation function and Mean Squared Error (MSE) as the loss function. The model is trained for up to 50 epochs.} Additionally, we incorporate early stopping, which continuously monitors the cross-validation set’s performance throughout the training process. If the validation error starts to increase, or if there is no improvement over a predetermined number of epochs, the training is stopped and the model with the best performance on the cross-validation set is selected as the final model.

We test and evaluate various LSTM-Autoencoder architectures and select \{100, 50, 50, 50, 100\} as it depicts the best performance in learning the characteristics of the training data and reconstructing them with low reconstruction errors. It is composed of two LSTM layers of 100 and 50 neurons, respectively, forming the encoder, and a decoder having the encoder’s mirrored architecture. The code of the LSTM-Autoencoder is 50 neurons. We train our model with the selected architecture after fine-tuning its hyperparameters.
%\vspace{-0.7cm}

\subsection {K-means architecture} 
\label{sec:K-means}
%We apply the K-means clustering algorithm in our approach, leveraging the default hyperparameters provided by the scikit-learn library \cite{scikit}. Specifically, we configured only the \texttt{random\_state} to a constant value and the number of clusters (\texttt{n\_clusters}), which was set to 2, as our approach addresses a binary classification problem using PUL. All other hyperparameters, such as the initialization method (\texttt{init}), the number of initializations (\texttt{n\_init}), the maximum number of iterations (\texttt{max\_iter}), and the convergence tolerance (\texttt{tol}), were retained at their default values, ensuring a standard and reproducible clustering process, as shown in Table \ref{tab:KmeanHyper}.

%\textcolor{blue}{We apply the K-means clustering algorithm in our approach, using the standard implementation provided by the scikit-learn library~\cite{scikit}. The parameter \texttt{n\_clusters} is set to 2 to reflect the binary classification objective, and \texttt{random\_state} is fixed at 42 to ensure reproducibility. The \texttt{init} parameter is set to \texttt{'k-means++'} to improve centroid initialization. We set \texttt{n\_init} to 10 to allow multiple initial centroid seeds, \texttt{max\_iter} to 300 to define the maximum number of iterations, and \texttt{tol} to \texttt{1e-4} as the convergence threshold. These settings ensure a stable and effective clustering process, enabling the model to separate benign and attack patterns in the latent feature space.}

\textcolor{black}{We apply the K-means clustering algorithm using the standard implementation provided by the scikit-learn library~\cite{scikit}. The number of clusters is set to two to align with the binary classification objective of our approach. To achieve consistent model behavior, the random seed is fixed to 42. The algorithm adopts the \enquote{k-means++} initialization method to enhance centroid selection and accelerate convergence.
To improve clustering robustness, the model is configured to run ten times with different centroid seeds. The maximum number of iterations allowed for convergence is defined as 300, and the tolerance threshold for convergence is specified as 1e-4. These settings ensure a stable and effective clustering process, enabling the model to accurately separate benign and attack patterns within the latent feature space extracted by the LSTM-Autoencoder.}
\vspace{-0.5cm}

\subsection {Benchmark Models} 
\label{sec:Benchmark}
%For benchmarking purposes, we implemented two custom PUL-based models: \textit{PUL-OCSVM-RF} and \textit{PUL-OCSVM-XGBoost}. Both models follow the PUL framework and share the same data preprocessing and feature extraction pipeline used in our main solution (\ref{sec:DataCollection},\ref{sec:fextrac}). %Specifically, they apply a variance threshold method to eliminate low-variance features and use \texttt{StandardScaler} for feature normalization.

%In accordance with the standard two-step PUL approach, both models first apply a One-Class Support Vector Machine (OCSVM) to the positively labeled benign samples in order to identify reliable negatives from the unlabeled set. The OCSVM employs an RBF kernel with \texttt{gamma='scale'}, and the \texttt{nu} parameter was set to 0.025 for \textit{PUL-OCSVM-RF} and 0.03 for \textit{PUL-OCSVM-XGBoost}. Unlabeled samples predicted as $-1$ (i.e., likely attacks) were selected as reliable negatives.

%These reliable negatives were then combined with the labeled benign data to train a supervised binary classifier. For \textit{PUL-OCSVM-RF}, a Random Forest classifier was used with \texttt{n\_estimators=100} and \texttt{random\_state=42}. For \textit{PUL-OCSVM-XGBoost}, we used an XGBoost classifier with \texttt{n\_estimators=100}, \texttt{max\_depth=5}, \texttt{learning\_rate=0.1}, \texttt{eval\_metric='logloss'}, \texttt{use\_label\_encoder=False}, and \texttt{random\_state=42}. These models were then evaluated on external test datasets to compare their performance with \textit{PUL-Inter-Slice Defender}.

\textcolor{black}{To conduct a thorough evaluation of our proposed solution, we compare the PUL-Inter-Slice Defender against three benchmark models. The first is the original Inter-Slice Defender \cite{rayalam2024}, which serves as a natural baseline since our PUL-based model builds upon its architectural foundation.}

\textcolor{black}{In addition, we implemented two custom PUL-based models: PUL-OCSVM-RF and PUL-OCSVM-XGBoost, to serve as comparative baselines within the PUL paradigm. Both models follow the same data pre-processing and feature extraction pipeline used in our main solution  (\sect{sec:DataCollection}, \sect{sec:fextrac}). }
% \vspace{-0.1cm}
\textcolor{black}{In accordance with the standard two-steps PUL approach \cite{bekker2020learning}, both models first apply OCSVM to the positively labeled benign samples to identify reliable negatives from the unlabeled set. OCSVM has strong capability in detecting outliers and anomalies in high-dimensional data \cite{agyemang2024anomaly}, making it particularly effective for identifying reliable negatives within the PUL context. It employs an Radial Basis Function (RBF) kernel with \texttt{gamma='scale'}. The \texttt{nu} parameter, which controls the trade-off between training errors and model complexity, was set to 0.025 for \textit{PUL-OCSVM-RF} and 0.03 for \textit{PUL-OCSVM-XGBoost}. Samples predicted as $-1$ (i.e., attacks) were selected as reliable negatives.}
\textcolor{black}{These reliable negatives were then combined with the labeled benign data to train a supervised binary classifier.} %For \textit{PUL-OCSVM-RF}, we used a Random Forest classifier with \texttt{n\_estimators=100} and \texttt{random\_state=42}. For \textit{PUL-OCSVM-XGBoost}, an XGBoost classifier was trained with \texttt{use\_label\_encoder=False}, \texttt{random\_state=42}, \texttt{n\_estimators=100}, \texttt{eval\_metric='logloss'}, \texttt{max\_depth=5}, and \texttt{learning\_rate=0.1}.}
  \textcolor{black}{For the PUL-OCSVM-RF model, we utilize a RF classifier, whose ensemble nature helps improve classification accuracy and robustness when working with labeled data \cite{awotunde2023multi}, making it particularly suitable for the second step of the two-step PUL technique \cite{bekker2020learning}. It was configured with 100 estimators and a fixed random seed of 42 to ensure reproducibility. For the PUL-OCSVM-XGBoost model, we train an XGBoost %\red{[HA]- provide some background on it, why you chose it, also add reference to it}
  classifier, selected for its demonstrated success in supervised anomaly detection and its ability to efficiently handle structured data , making it well suited for classification tasks within the PUL setting \cite{dilworth2024harnessing}. The model is configured with the label encoder disabled, a fixed random seed of 42, an evaluation metric based on log loss, 100 estimators, a maximum tree depth of 5, and a learning rate of 0.1.}
  % with the label encoder disabled, a fixed random seed of 42, and an evaluation metric based on log loss. The model also uses 100 estimators, a maximum tree depth of 5, and a learning rate of 0.1.}

\textcolor{black}{ These three benchmark models (Inter-Slice Defender, PUL-OCSVM-RF, and PUL-OCSVM-XGBoost) were trained and evaluated using the same datasets as the PUL-Inter-Slice Defender (Section \ref{sec:dataset}), including training sets with varying levels of contamination and a common set of test datasets. The corresponding evaluation results are presented in the following subsection.}

\vspace{-0.3cm}
\subsection {PUL-Inter-Slice Defender Performance} 
\label{sec:ISdefender}
   
%To evaluate the performance and robustness of our proposed approach, PUL-Inter-Slice Defender, we conduct a comparative analysis against Inter-Slice Defender \cite{rayalam2024} \textcolor{red}{when trained using contaminated data.} We retrain Inter-Slice Defender using our \textcolor{red}{four training datasets (Section VI-C) and test its performance after training with each dataset to evaluate its performance under different levels of contamination (i.e., 10\%, 20\%, 30\%, 40\%) in the training data.}  

%To conduct a thorough evaluation of our proposed solution, we compare the PUL-Inter-Slice Defender against  three reference models: Inter-Slice Defender, PUL-OCSVM-RF and PUL-OCSVM-XGBoost

%This comparison is carried out across multiple contamination levels in the training dataset (Table \ref{tab:DataForTests}), aiming to highlight each model’s capability to maintain reliable detection performance under realistic conditions where the presence of mislabeled or anomalous data is common }

We now present the performance of the proposed PUL-Inter-Slice Defender and compare it against the benchmark models (Section \ref{sec:Benchmark}) across varying contamination levels.

%We retrain Inter-Slice Defender using each of our four training datasets (Section \ref{sec:dataset}) to evaluate its performance under different levels of contamination (i.e., 10\%, 20\%, 30\%, 40\%). The model’s lookback hyperparameter (i.e., timesteps) is fixed at 1, and the detection threshold is set to 0.1408, based on findings from \cite{rayalam2024}, which identified these values as optimal for detection performance. Figure \ref{fig:RSA_TSA} shows a significant decline in detection performance for RSA and TSA under contaminated data conditions, with TSA more affected than RSA. TSA’s F1-score drops from 76.46\% to 44.36\%, and RSA’s from 85.01\% to 61.86\% as contamination levels increase from 10\% to 40\%. This decline is due to the model’s design, which assumes training on a clean dataset (i.e., entirely benign), a condition that is often not met in real-world deployments where contaminated data is common. Contrary to this assumption, real-world deployments frequently contend with a significant presence of contaminated data, which undermines the effectiveness of models in practice.

We retrain Inter-Slice Defender using each of our four training datasets (Section \ref{sec:dataset}) to evaluate its performance under different levels of contamination (i.e., 10\%, 20\%, 30\%, 40\%). The model’s lookback hyperparameter (i.e., timesteps) is fixed at 1, and the detection threshold is set to 0.1408, based on findings from \cite{rayalam2024}, which identified these values as optimal for detection performance. Fig. \ref{fig:averages} shows Inter-Slice Defender's F1-score values ranging from 80.74\% to 53.11\% as contamination levels increase from 10\% to 40\% for detecting RSA and TSA. \textcolor{black}{This performance drop is attributed to the model’s assumption of a clean training dataset (i.e., entirely benign), an assumption that does not hold in real-world deployments, where contamination is frequently present and compromises the model’s practical effectiveness.}

%This decline is due to the model’s design, which assumes training on a clean dataset (i.e., entirely benign), a condition that is often not met in real-world deployments, where contaminated data is common. Contrary to this assumption, real-world scenarios frequently contend with a significant presence of contaminated data, which undermines the practical effectiveness of such models

%\textcolor{blue}{The PUL-XGBoost (Figure \ref{fig:P_XGBoosT}) and PUL-RF (Figure \ref{fig:P_RF}) models both acknowledge the realities of real-world deployments, where training data often includes mislabeled or anomalous samples (e.g., contaminated dataset). By integrating the PUL framework, both models exhibit improved detection performance compared to the baseline Inter-Slice Defender (\ref{fig:RSA_TSA}), which assumes clean training data. As illustrated in Figures \ref{fig:P_XGBoosT} and \ref{fig:P_RF}, both PUL-XGBoost and PUL-RF achieve higher F1-scores than Inter-Slice Defender under all contamination levels (10\%, 20\%, 30\%, and 40\%), highlighting their relative improvement due to the integration of PUL. Specifically, PUL-XGBoost maintains an RSA F1-score of 93.24\% at 10\% contamination, which gradually decreases to 75.39\% at 40\%, while TSA detection drops from 86.49\% to 62.06\%. Similarly, PUL-RF records an RSA F1-score ranging from 91.83\% to 75.91\%, and TSA performance declines from 82.58\% to 61.29\% over the same contamination range.}

\textcolor{black}{In contrast, the performance of PUL-OCSVM-XGBoost and PUL-OCSVM-RF reveals a clear advantage compared to the Inter-Slice Defender across all contamination levels (10\%, 20\%, 30\%, and 40\%), as shown in Fig. \ref{fig:averages}, highlighting the benefits of PUL.  Specifically, PUL-OCSVM-XGBoost maintains an F1-score of 89.87\% at 10\% contamination, which gradually decreases to 68.73\%  as the contamination level reaches 40\%. Similarly, PUL-OCSVM-RF records an F1-score ranging from 87.21\% to 68.60\% over the same contamination range.}
\textcolor{black}{Despite these improvements, both PUL-OCSVM-XGBoost and PUL-OCSVM-RF models exhibit a noticeable decline in performance as contamination increases, reflecting their limited capacity to capture temporal dependencies in the data. This poses a critical limitation for DSM attack detection, where RSA and TSA unfold through sequences of 5G control procedures such as registration, authentication, and PDU session establishment. These procedures generate time series patterns essential for distinguishing anomalous from benign behavior. While RF treats each input independently and is not tailored for capturing temporal dependencies; XGBoost does not account for sequential dependencies like LSTM. This limitation can lead to misclassifications when subtle temporal patterns become decisive under high contamination conditions.}

\textcolor{black}{On the contrary, PUL-Inter-Slice Defender demonstrates superior performance in detecting both RSA and TSA attacks, as illustrated in Fig. \ref{fig:averages}. It achieves F1-scores ranging from 99.34\% to 98.50\% across training datasets with contamination levels varying between 10\% and 40\%, significantly outperforming the benchmark models: Inter-Slice Defender, PUL-OCSVM-XGBoost, and PUL-OCSVM-RF. This performance improvement is attributed to three key architectural strengths:}

%In contrast, PUL-Inter-Slice Defender demonstrates superior performance in detecting both RSA and TSA attacks, as illustrated in Fig. \ref{fig:averages}. It achieves F1-scores ranging from 99.34\% to 98.50\% across contamination rates of 10\% to 40\% in the training dataset. These results substantially exceed the performance of the benchmark models: Inter Slice Defender, PUL-OCSVM-XGBoost, and PUL-OCSVM-RF, which underscore the robustness of our approach in the presence of varying levels of data contamination. This significant improvement can be attributed to three key architectural strengths of the proposed model.

%\textcolor{blue}{First, the PUL-Inter-Slice Defender incorporates an LSTM-autoencoder for feature extraction to effectively handle the sequential nature of the data. In the context of DSM attacks such as RSA and TSA, the behavior of the system unfolds over time through a series of control plane procedures. These procedures naturally generate time series data, where the order and timing of events are critical to identifying abnormal patterns. The LSTM autoencoder is well suited for this task, as it captures temporal dependencies and encodes them into a compact and informative representation of the input sequences.}
\begin{enumerate}
 \item \textcolor{black}{The integration of LSTM-autoencoder to effectively handle the sequential nature of the data. In the context of DSM attacks such as RSA and TSA, the behavior of the system unfolds over time through a series of control plane procedures. These procedures naturally generate time series data, where the order and timing of events are critical for identifying normal and abnormal patterns.}
%The integration of LSTM-autoencoder to capture temporal patterns in CP procedures, enabling effective detection of DSM attacks like RSA and TSA.

%\textcolor{blue}{Second, the model leverages the latent space produced by the LSTM-autoencoder to enhance its detection capabilities. This latent representation captures essential temporal dependencies and interrelations among 5G control plane procedures that are often critical for distinguishing benign and malicious behaviors. It preserves the sequential structure of the original data while filtering out irrelevant variability, allowing the model to focus on patterns most indicative of benign behavior, RSA and TSA. The use of the latent space brings several advantages: it reduces noise, preserves meaningful structure, and enables the model to focus on the most relevant and informative patterns. These properties make it particularly effective for the clustering process in our framework, as they help form clear separations between benign and attack behaviors, even when the training data contains high levels of contamination.}

\item \textcolor{black}{The model leverages the LSTM-autoencoder’s latent space to extract abstract features, facilitating effective feature clustering and pattern separation.}
%Second, the model leverages the LSTM-autoencoder’s latent space to capture temporal dependencies and key structural patterns in 5G control plane procedures. By preserving sequence integrity and reducing noise, it highlights features critical to distinguishing benign behavior from RSA and TSA, enabling effective clustering.

%\textcolor{blue}{Third, the integration of the PUL framework further improves the model’s adaptability to real world conditions. In realistic deployment scenarios, as mentioned earlier, obtaining fully labeled datasets is often infeasible, and contamination in the training data is common. The PUL approach enables the model to learn robust decision boundaries from a small subset of positively labeled samples and a larger set of unlabeled data. As a result, the model is particularly effective in dynamic environments where only limited labeled data is available, and uncertainty in the training distribution poses a significant challenge to accurate detection.}

\item \textcolor{black}{PUL enhances adaptability by enabling robust decision boundary learning from limited labeled positives and a large pool of unlabeled samples, making it well-suited for real-world scenarios with contaminated training data.}
\end{enumerate}

%\textcolor{blue}{Overall, by combining the strengths of PUL learning and temporal feature modeling via the latent space derived from the LSTM autoencoder, the PUL-Inter-Slice Defender achieves exceptional detection performance in challenging conditions where training data is contaminated. This synergy enables the model to generalize well despite label uncertainty and temporal complexity, making it a robust and practical solution for real-world DSM attack detection in 5G networks.} 

\textcolor{black}{Overall, by combining PUL with the LSTM-autoencoder’s latent space, the PUL-Inter-Slice Defender effectively captures the sequential structure and timing of 5G CP procedures. This enables robust detection under contaminated training conditions and allows the model to generalize well despite label uncertainty and the complexity of time-dependent behaviors during a DSM attack.}

\section{\textcolor{black}{Implementation Considerations in Real-World 5G Environment}} \label{sec:deploy}

\textcolor{black}{PUL-Inter-Slice Defender leverages 3GPP KPIs and PM counters as features for DSM attack detection. These metrics are usually calculated by different 5G NFs and shared with the Network Data Analytics Function (NWDAF)}%This capability can be effectively supported through its integration with
\textcolor{black}{, which is a core 5G NF designed to collect, process, and analyze data across various network components \cite{nwdaf}. NWDAF provides a granular view of network behavior and can be configured to monitor targeted NSs and NFs such as AMF, SMF, and NSSF, which are critical components for maintaining the operational integrity and security of the 5G network. By leveraging NWDAF’s data collection and analytics capabilities, PUL-Inter-Slice Defender enhances its effectiveness in detecting DSM attacks, offering a data-driven approach that strengthens the overall resilience of the 5G network. Additionally, PUL-Inter-Slice Defender can complement existing intrusion detection systems to trigger automated responses such as rate limiting or slice isolation, mitigating threats before they escalate.}

\textcolor{black}{However, deploying PUL-Inter-Slice Defender in real-world 5G networks presents several challenges, particularly due to the fragmented nature of NS deployments across multiple administrative domains. These domains often enforce distinct security policies and data-sharing restrictions, limiting the holistic analysis of network behavior. Privacy concerns further complicate collaboration, as operators may be unwilling to share sensitive operational data. These limitations hinder centralized anomaly detection and necessitate approaches that balance accuracy, efficiency, and data privacy.}

\textcolor{black}{To address these constraints, PUL-Inter-Slice Defender's integration with NWDAF should support distributed monitoring, enabling each domain to maintain autonomy while contributing to a collective detection framework. A promising direction for future work is the use of federated learning \cite{wen2023survey}, where local models are trained within each domain and only aggregated insights are shared. This allows for collaborative, privacy-preserving anomaly detection that is scalable, efficient, and aligned with the operational constraints of diverse 5G environments.}
\vspace{-0.3cm}

\section {Conclusion} \label{sec:conclussi}
%\vspace{-0.3cm}
In this work, we introduced, tested, and evaluated two variants of the known DSM attack, namely RSA and TSA, that exploit ISS procedures. We analyzed the impact that these attacks have on the 5G network using our free5GC testbed and UERANSIM simulator. We showed that they cause a DDoS on the network due to the resulting overload on the AMF.
Furthermore, we developed the PUL-Inter-Slice Defender, an innovative anomaly detection solution designed to identify inter-slice attacks such as RSA and TSA  in the presence of contaminated data. PUL-Inter-Slice Defender employs the PUL approach, integrating LSTM-Autoencoder with K-Means. The proposed solution leverages 3GPP KPIs and PM counters for its training process. The integration of these 3GPP features enables easy deployment of our PUL-Inter-Slice Defender as part of the NWDAF, as these features are usually made/can be available in this 5G NF. %PUL-Inter-Slice Defender was rigorously evaluated under various conditions, including varying architectures, and levels of dataset contamination, as well as across different test datasets, demonstrating the framework’s robust capability to detect RSA and TSA. Notably, it achieved an average F1-score exceeding 98.50\% across different training datasets containing various percentages of negative samples.
PUL-Inter-Slice Defender was rigorously evaluated under various conditions, including different architectures, contamination levels, and test datasets. \textcolor{black}{It was also compared against three benchmark models (Inter-Slice Defender, PUL-OCSVM-RF, and PUL-OCSVM-XGBoost), outperforming them and achieving an average F1-score exceeding 98.50\% across training datasets with varying proportions of negative samples, highlighting its robustness and generalization capability in detecting RSA and TSA.}

Finally, as a future work, we aim at studying the performance and economic impact that the DSM attack and its variations can have on NSs when the AMF and CP NFs resources are large enough to contain the attack. \textcolor{black}{Additionally, we plan to investigate the impact of concept drift, particularly under dynamic and evolving 5G network conditions where the statistical distribution of RSA, TSA, or benign traffic may change over time.}

\vspace{-0.3cm}
%\section*{Acknowledgments}

%This research is made possible through the financial support of Concordia University, Ericsson research, Montreal, and a grant from the National Cyber security Consortium Canada under the Cyber Security Innovation Network.
%\vspace{-0.5cm}

\bibliography{refs}
\bibliographystyle{IEEEtran}
%\printbibliography
%\input{bibliography}

\end{document}